\documentclass[journal]{IEEEtran}
\usepackage{amsmath,amsfonts}
\usepackage{booktabs,makecell,tabularx}
\usepackage{algorithm}

\usepackage{array}
\usepackage[caption=false,font=normalsize,labelfont=sf,textfont=sf]{subfig}
\usepackage{textcomp}
\usepackage{stfloats}
\usepackage{url}
\usepackage{verbatim}
\usepackage{graphicx}
\usepackage[algo2e]{algorithm2e}
\usepackage{cite}
\usepackage{xcolor}
\usepackage{hyperref}
\usepackage{float}
\usepackage[table]{xcolor}

\def\BibTeX{{\rm B\kern-.05em{\sc i\kern-.025em b}\kern-.08em
    T\kern-.1667em\lower.7ex\hbox{E}\kern-.125emX}}
\usepackage{balance}
\begin{document}
\title{DSXFormer: Dual-Pooling Spectral Squeeze-Expansion and Dynamic Context Attention Transformer for Hyperspectral Image Classification}
\author{ Farhan Ullah, Irfan Ullah, Khalil Khan, Giovanni Pau, JaKeoung Koo
\thanks{Corresponding author: JaKeoung Koo
\par Farhan Ullah and JaKeoung Koo are with the School of Computing, Gachon University, Seongnam 13120, South Korea. (farhan.marwat@gmail.com, jakeoung@gachon.ac.kr)
\par Irfan Ullah is with the School of Computer Science, Chengdu University of Technology, Sichuan, China (irfan.ee@cqu.edu.cn)
\par Khalil Khan is with the Department of Information Technology, College of Computer, Qassim University, Buraydah, Saudi Arabia (k.sirkhan@qu.edu.sa)
\par Giovanni Pau is with the Kore University of Enna, Italy (giovanni.pau@unikore.it)}}

\markboth{}%
{DSXFormer: Dual-Pooling Spectral Squeeze-Expansion and Dynamic Context Attention Transformer for Hyperspectral Image Classification}

\maketitle

\begin{abstract}
Hyperspectral image classification (HSIC) is a challenging task due to high spectral dimensionality, complex spectral–spatial correlations, and limited labeled training samples. Although transformer-based models have shown strong potential for HSIC, existing approaches often struggle to achieve sufficient spectral discriminability while maintaining computational efficiency. To address these limitations, we propose a novel DSXFormer, a novel dual-pooling spectral squeeze-expansion transformer with Dynamic Context Attention for HSIC. The proposed DSXFormer introduces a Dual-Pooling Spectral Squeeze-Expansion (DSX) block, which exploits complementary global average and max pooling to adaptively recalibrate spectral feature channels, thereby enhancing spectral discriminability and inter-band dependency modeling. In addition, DSXFormer incorporates a Dynamic Context Attention (DCA) mechanism within a window-based transformer architecture to dynamically capture local spectral–spatial relationships while significantly reducing computational overhead. The joint integration of spectral dual-pooling squeeze–expansion and DCA enables DSXFormer to achieve an effective balance between spectral emphasis and spatial contextual representation. Furthermore, patch extraction, embedding, and patch merging strategies are employed to facilitate efficient multi-scale feature learning. Extensive experiments conducted on four widely used hyperspectral benchmark datasets, including Salinas (SA), Indian Pines (IP), Pavia University (PU), and Kennedy Space Center (KSC), demonstrate that DSXFormer consistently outperforms state-of-the-art methods, achieving classification accuracies of 99.95\%, 98.91\%, 99.85\%, and 98.52\%, respectively.

\end{abstract}


\section{Introduction}
\IEEEPARstart{H}YPERSPECTRAL images (HSIs) acquire comprehensive spectral information of ground objects by capturing hundreds of contiguous spectral bands. This high spectral resolution provides significantly enhanced capability for discriminating land-cover classes compared with traditional color or multispectral imagery \cite{1huang2021subspace,2wang2023penetrating}. As a result, hyperspectral imaging has been widely adopted in numerous application domains, including mineral exploration \cite{3peyghambari2021hyperspectral}, environmental monitoring \cite{4stuart2019hyperspectral}, and medical diagnosis \cite{5lv2021spatial}. In hyperspectral image analysis, classification remains a core and indispensable task for interpreting the rich spectral–spatial information embedded in HSIs.

Conventional HSIC approaches predominantly employed traditional machine learning classifiers, including k-nearest neighbors (KNN) \cite{6blanzieri2008nearest}, support vector machines (SVMs) \cite{7melgani2004classification}, random forests (RFs) \cite{8zhang2018cascaded}, and sparse representation-based classification (SRC) \cite{9chen2010classification,10peng2018maximum}. The performance of these methods is largely contingent upon the quality of manually engineered features. Depending on the type of features utilized, they are typically categorized as spectral-based \cite{11xia2017hyperspectral,12li2011locality} or spectral–spatial-based approaches \cite{13zhang2017multifeature,14ren2017hyperspectral}.

Spectral-based methods perform classification by directly exploiting the spectral signatures of HSIs, either using the original spectral vectors or their transformed representations, thereby considering only spectral information\cite{52Ullah,53ullah}. However, the inherently high dimensionality of hyperspectral data often leads to the curse of dimensionality when raw spectral features are used directly. To alleviate this issue, feature transformation-based methods typically apply dimensionality reduction through feature extraction or feature selection techniques \cite{16huang2022heterogeneous,17huang2021structural}, such as principal component analysis (PCA) \cite{18camps2006composite}, linear discriminant analysis (LDA) \cite{19sadeghi2021neural}, and band selection, followed by classification in a reduced-dimensional feature space. These transformed features generally exhibit stronger discriminative capability, resulting in improved classification performance compared with raw spectral features. Nevertheless, because of the lack of explicit spatial context, spectral-based methods remain highly susceptible to noise and outliers, which can adversely affect classification accuracy.

To address these limitations, spectral–spatial-based methods have been developed to jointly exploit both spectral and spatial information in HSIs. Prominent approaches can be broadly categorized into pre-processing-based \cite{20jia2017local} and post-processing-based methods \cite{21liu2015probabilistic}. Pre-processing-based techniques focus on extracting robust spectral–spatial features prior to classification, using methods such as Gabor filtering \cite{22fauvel2009kernel} and morphological profiles \cite{23fauvel2008spectral}. In contrast, post-processing-based approaches aim to enhance spatial consistency by applying filtering operations to the initial classification maps \cite{24zhao2022superpixel}. Although these strategies substantially outperform purely spectral-based methods, conventional HSIC techniques generally extract only shallow-level features, thereby limiting their ability to effectively model the complex and highly nonlinear structures inherent in hyperspectral data.

In recent years, deep learning techniques have achieved remarkable success in computer vision and the analysis of high-dimensional data \cite{59ullah,58khan,60ullah}. In contrast to conventional handcrafted feature extraction approaches, deep learning-based methods are capable of automatically learning hierarchical and discriminative representations directly from data, thereby significantly enhancing performance across a wide range of application domains \cite{25liu2024crossmatch,61ullahlarge,62ullahvisual,63khan2025role}. Consequently, numerous deep learning-based frameworks have been introduced for HSIC, employing various neural network architectures, such as recurrent neural networks (RNNs) \cite{26liang2022multiscale}, generative adversarial networks (GANs) \cite{27zhang2020optimized}, and graph convolutional networks (GCNs) \cite{28liu2020cnn}. These approaches have demonstrated substantial performance gains over traditional shallow classifiers.

Among existing architectures, convolutional neural networks (CNNs) have emerged as the most extensively adopted models for HSIC, owing to their computational efficiency and strong capability to capture local spatial structures. Representative CNN-based models include one-dimensional CNNs (1D-CNNs) \cite{29hu2015deep}, two-dimensional CNNs (2D-CNNs) \cite{30makantasis2015deep}, and three-dimensional CNNs (3D-CNNs) \cite{31chen2016deep,54Ullah}, which are designed to extract spectral features, spatial features, and joint spectral–spatial features, respectively. To further enhance representation learning, hybrid architectures that integrate 2D-CNN and 3D-CNN have been proposed, achieving superior performance compared with standalone 3D-CNN models \cite{32roy2019hybridsn}.

In addition, to facilitate the extraction of deeper and more robust features, skip-connected CNN architectures, such as ResNet \cite{33zhong2017spectral} and DenseNet \cite{34zhang2019multi}, have been introduced for HSI classification, effectively alleviating information degradation across network layers. Furthermore, Dong et al. \cite{35dong2022deep} incorporate grey-level co-occurrence matrices to strengthen spatial feature modeling. Inspired by the attention mechanism, a series of attention-enhanced CNN-based networks have also been developed \cite{36sun2019spectral,37lu20203}. For instance, Zhu et al. \cite{39zhu2020residual} jointly employ spectral and spatial attention mechanisms to adaptively reweight feature responses across both channel and spatial dimensions, enabling the network to focus on more informative features for improved HSIC.

While CNN-based classification models demonstrate strong empirical performance, they are inherently constrained by the limited receptive field of convolutional operations, particularly when small kernel sizes are employed, which restricts their ability to effectively model long-range dependencies. In contrast, vision transformers (ViTs) \cite{40dosovitskiy2020image} have recently achieved notable success in hyperspectral image (HSI) classification. Leveraging the self-attention (SA) mechanism \cite{41vaswani2017attention}, ViTs substantially expand the effective receptive field relative to CNNs, thereby enabling more effective modeling of global contextual relationships within hyperspectral data.

A number of transformer-based approaches for HSIC have been proposed \cite{42xie2023semantic,43mei2022hyperspectral,44Zhao}, aiming to enhance the standard ViT architecture by improving tokenization strategies or refining attention mechanisms. For example, Hong et al. \cite{48hong2021spectralformer} construct input tokens by grouping neighboring spectral bands prior to encoding, while FUST \cite{49zeng2023microscopic} directly feeds flattened HSI patches into a transformer branch to capture long-range spectral dependencies. Rather than generating tokens solely from raw hyperspectral data, many recent studies have adopted hybrid CNN–transformer architectures to more effectively exploit complementary spatial–spectral characteristics. In this context, Sun et al. \cite{50sun2022spectral} employ a CNN-based module for feature extraction and token generation, an approach that has been subsequently extended in \cite{51yu2022mstnet,52ghosh2022hyperspectral}. More recently, hierarchical and specialized transformer architectures, including Spectral Swin Transformer (SwinT) \cite{47liu2023spectral}, pyramid-based PyFormer \cite{48pyramid}, multi-path context modeling networks such as PMCN \cite{51PMCN}, frequency-aware WaveFormer \cite{50waveformer}, and ViT-based hyperspectral variants \cite{46ayas2022spectralswin}, have further improved global context modeling for HSIC.

Furthermore, HiT \cite{55yang2022hyperspectral} enhances local spatial and spectral feature extraction by replacing the SA blocks in the transformer with convolution-based operations, referred to as a convolutional permutator. While this design leads to improved classification performance, its capacity to model long-range dependencies across both spatial and spectral dimensions remains limited. To address this issue, the methods proposed in \cite{56zhang2022convolution,57xu2023multiscale} incorporate convolutional operations within the multi-head self-attention (MSA) blocks of the transformer encoder, thereby strengthening local feature modeling while retaining global attention capabilities. Additionally, recent studies \cite{58yang2023gtfn,59Jiang} incorporate graph-based representations into transformer frameworks to more effectively encode spatial–spectral relationships during tokenization, facilitating improved local-to-global feature learning. In particular, GraphGST \cite{59Jiang} explicitly captures the physical positional relationships among pixels by constructing positional encodings based on the frequency distribution of patch labels within a sliding window.

Although the aforementioned approaches have demonstrated strong performance in HSIC, several challenges remain unresolved. First, the self-attention (SA) mechanism employed in vision transformer (ViT) encoders is primarily effective at capturing global contextual information, but it exhibits limited capability in modeling fine-grained local features, particularly in scenarios with insufficient training samples. Second, the substantial number of parameters introduced by complex architectural designs during tokenization and feature extraction within transformer encoders increases the susceptibility to overfitting when training data are scarce. Although recent lightweight transformer-based models have been proposed \cite{60zhao2024hyperspectral}, their ability to adequately represent HSIs with highly complex and heterogeneous structures remains insufficient.

To address these limitations, this paper proposes a novel DSXFormer method for HSIC. The proposed framework introduces a dual pooling–based spectral squeeze–expansion (DSX) module into a hierarchical transformer architecture, enabling effective joint modeling of fine-grained spectral dependencies and long-range spatial–spectral contextual information. The DSX module operates by first aggregating complementary global and local spectral statistics through dual pooling strategies, followed by a spectral expansion and recalibrating process that adaptively emphasizes informative spectral bands while suppressing redundant responses. This mechanism facilitates the learning of discriminative spectral representations with enhanced robustness under limited training samples.

Furthermore, DSXFormer employs a window-based DCA mechanism to efficiently capture long-range dependencies while maintaining computational efficiency. By integrating the DSX module within the transformer encoder, the proposed architecture effectively balances global contextual modeling and local feature refinement, alleviating the limitations of conventional ViT-based methods. Through adaptive spectral recalibration and hierarchical contextual aggregation, DSXFormer significantly improves classification accuracy and generalization capability, providing a robust and efficient solution for HSIC. The main contributions of our paper are summarized as follows:

\begin{enumerate}
\item We propose DSXFormer, a novel transformer-based framework for HSIC that integrates spectral dual-pooling squeeze-expansion and attention mechanisms to jointly capture spectral and spatial dependencies with high precision.
\item We introduce a Dual-Pooling Spectral Squeeze-Expansion (DSX) block, which leverages complementary global average and max pooling to enhance spectral feature representation while maintaining computational efficiency.
\item We design a DCA mechanism that incorporates relative positional encoding and similarity-guided context scaling, enabling more discriminative and adaptive feature learning within local windows.
\item Extensive experiments on multiple benchmark hyperspectral datasets demonstrate that DSXFormer consistently outperforms state-of-the-art transformer-based and CNN-based methods in terms of classification accuracy, robustness under limited training samples, and computational efficiency.
\end{enumerate}

The remainder of this paper is organized as follows. Section II introduces the proposed DSXFormer framework, detailing its overall architecture and key design components. Section III describes the experimental setup and presents a comprehensive evaluation of the proposed method on four widely used HSIC benchmarks. Section IV presents the paper's conclusion and discusses potential directions for future research.

\section{Proposed Method}
\begin{figure*}[t]
\centering
\includegraphics[width=\linewidth]{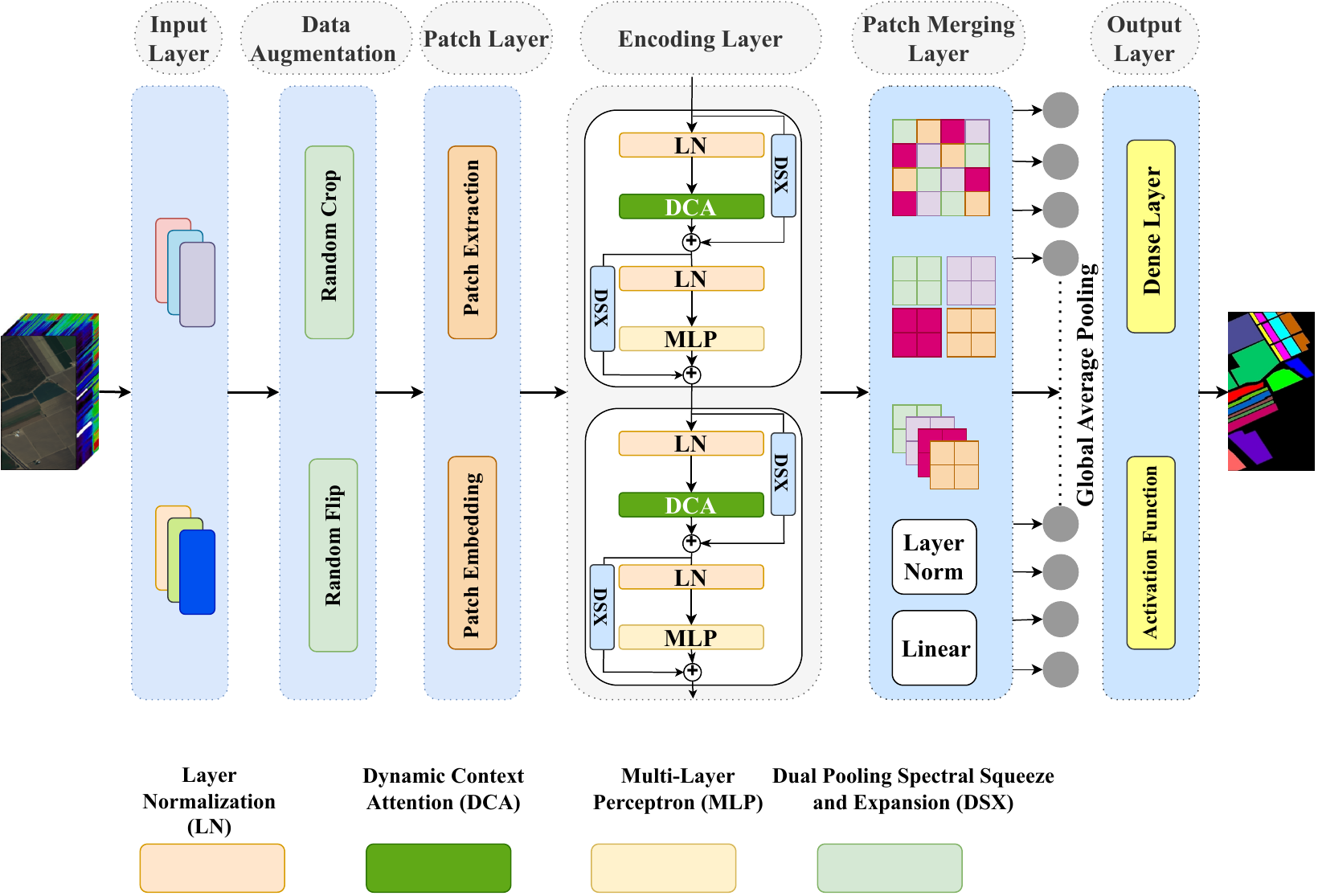}
\caption{Overall architecture of the proposed DSXFormer framework for HSIC. The network comprises patch extraction and embedding, hierarchical DSXFormer encoding blocks integrating DSX and DCA, followed by patch merging and a global pooling–based classification head as the output layer.}
\label{fig: Main}
\end{figure*}
\subsection{DSXFormer: Dual-Pooling Spectral Squeeze-Expansion Transformer Architecture}
In this section, we present our proposed DSXFormer, a Dual-Pooling Spectral Squeeze–Expansion Transformer that integrates context-aware self-attention mechanisms to enhance HSIC by effectively capturing discriminative spectral–spatial dependencies while maintaining computational efficiency. The overall architecture of the proposed DSXFormer is illustrated in Fig. \ref{fig: Main}, and the detailed components of the framework are described step by step in the following subsections.

\subsection{Overview of the Proposed DSXFormer Framework}
Let $\mathbf{H} \in \mathbb{R}^{R \times C \times L}$ denote a HSI with spatial dimensions \( R \) (height) and \( C \) (width), and \( L \) spectral bands. The objective of HSIC is to assign each pixel $\mathbf{h}_{r,c} \in \mathbb{R}^{L}$ to one of \( K \) predefined land-cover classes, resulting in a classification map $\mathbf{Y} \in \{1,\dots,K\}^{R \times C}$.

To this end, the input HSI $\mathbf{H}$ is first partitioned into non-overlapping spatial patches. Each patch is flattened and linearly projected into a latent embedding space, forming a sequence of patch embeddings \( Z \in \mathbb{R}^{N \times d} \), where \( N \) denotes the number of patches and \( d \) is the embedding dimension. To enhance spectral discriminability at an early stage, the proposed Dual-Pooling Spectral Squeeze--Expansion (DSX) block is incorporated to recalibrate channel-wise spectral responses by selectively emphasizing informative spectral bands within each patch.

The refined patch embeddings are then processed by a hierarchical transformer encoder equipped with window-based DCA, which facilitates localized feature interactions within fixed-size \( M \times M \) windows. To further capture cross-window dependencies and improve global contextual modeling, a shifted window strategy is employed, enabling effective information exchange between adjacent windows without incurring high computational cost. At successive stages, patch merging operations are applied to progressively reduce spatial resolution while increasing feature dimensionality, thereby supporting efficient multi-scale spectral--spatial representation learning.

Finally, the high-level feature representations are fed into a fully connected classification head to predict class labels for each patch. When required, the patch-level predictions are upsampled to recover a dense pixel-wise classification map \( Y \) at the original spatial resolution. By tightly integrating dual-pooling spectral recalibration and DCA transformer encoding, the proposed DSXFormer framework effectively balances classification accuracy and computational efficiency for HSIC.

\subsection{Dual-Pooling Spectral Squeeze-Expansion (DSX) Block}

\begin{figure*}[t]
\centering
\includegraphics[width=\linewidth]{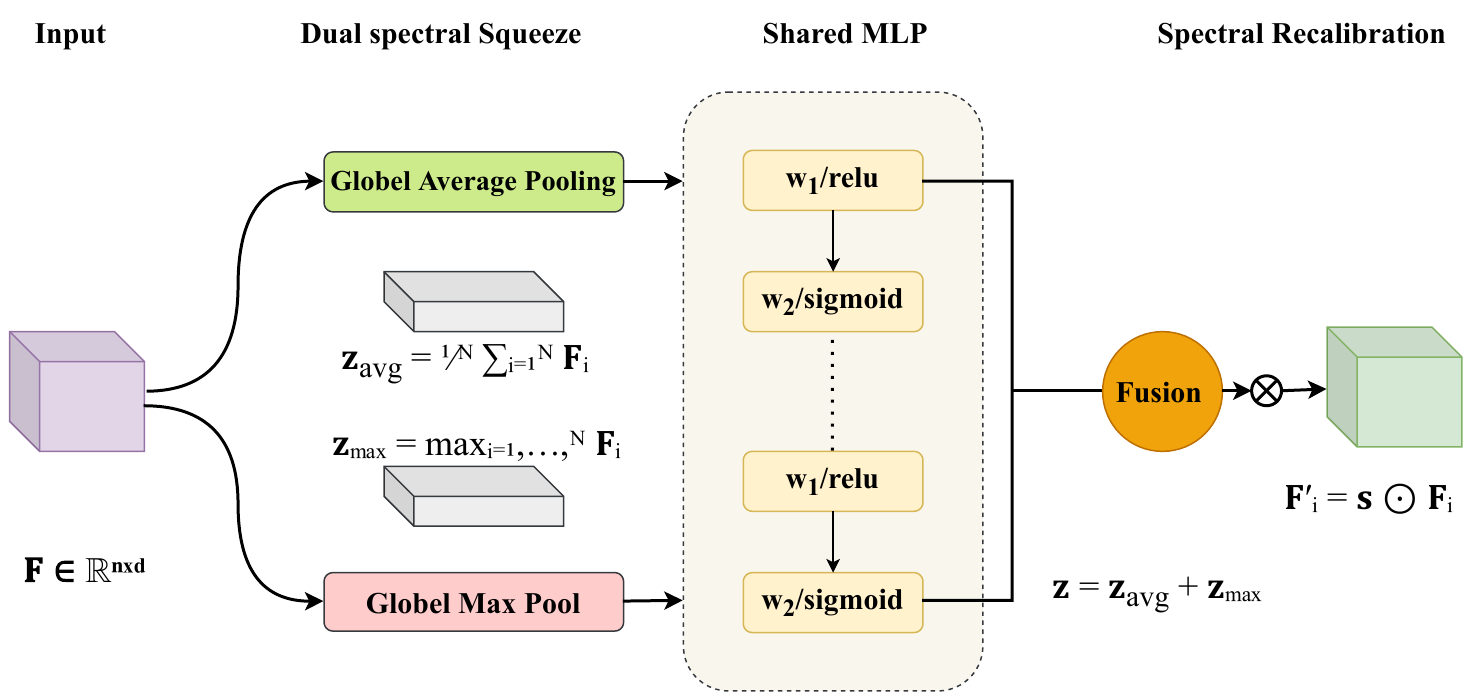}
\caption{Architecture of the proposed DSX module.}
\label{fig: SX_block}
\end{figure*}

Hyperspectral images are characterized by high spectral dimensionality, where different spectral bands contribute unequally to class discrimination. Efficiently modeling inter-band dependencies while suppressing redundant spectral responses is therefore crucial for high-precision HSIC. To this end, we propose the DSX block, a lightweight yet effective spectral recalibration module that adaptively emphasizes informative spectral channels and enhances spectral discriminability within transformer-based architectures. Let the input feature map to the DSX block be denoted as:
\begin{equation}
\mathbf{F} \in \mathbb{R}^{N \times d},
\end{equation}
where \( N \) represents the number of tokens (patches) and \( d \) denotes the embedding dimension corresponding to spectral feature channels. The DSX block operates in three main stages: dual-pooling spectral squeeze, spectral expansion and compression, and spectral feature recalibration.

\subsubsection{Dual-Pooling Spectral Squeeze}

To capture complementary global spectral statistics, the DSX block employs both \emph{global average pooling} (GAP) and \emph{global max pooling} (GMP) across the token dimension. Specifically, the pooled spectral descriptors are computed as:
\begin{equation}
\mathbf{z}_{\text{avg}} = \frac{1}{N} \sum_{i=1}^{N} \mathbf{F}_i,
\end{equation}
\begin{equation}
\mathbf{z}_{\text{max}} = \max_{i=1,\dots,N} \mathbf{F}_i,
\end{equation}
where \( \mathbf{F}_i \in \mathbb{R}^{d} \) denotes the feature vector of the \( i \)-th token. The average pooling operation captures the overall spectral distribution, while max pooling highlights the most salient spectral activations. These complementary descriptors are combined to form a unified spectral representation:
\begin{equation}
\mathbf{z} = \mathbf{z}_{\text{avg}} + \mathbf{z}_{\text{max}}.
\end{equation}

\subsubsection{Spectral Expansion and Compression}

The aggregated spectral descriptor \( \mathbf{z} \) is passed through a lightweight gating network composed of two fully connected layers to model nonlinear inter-band dependencies. First, the spectral dimension is expanded by a factor \( r \) to increase representational capacity:
\begin{equation}
\mathbf{h} = \delta \left( \mathbf{W}_1 \mathbf{z} + \mathbf{b}_1 \right),
\end{equation}
where \( \mathbf{W}_1 \in \mathbb{R}^{rd \times d} \), \( \mathbf{b}_1 \in \mathbb{R}^{rd} \), \( r \) denotes the expansion ratio, and \( \delta(\cdot) \) represents a nonlinear activation function (ReLU). Subsequently, the expanded features are compressed back to the original embedding dimension as:
\begin{equation}
\mathbf{s} = \sigma \left( \mathbf{W}_2 \mathbf{h} + \mathbf{b}_2 \right),
\end{equation}
where \( \mathbf{W}_2 \in \mathbb{R}^{d \times rd} \), \( \mathbf{b}_2 \in \mathbb{R}^{d} \), and \( \sigma(\cdot) \) denotes the sigmoid activation function. The resulting vector \( \mathbf{s} \in \mathbb{R}^{d} \) serves as a set of spectral attention weights.

\subsubsection{Spectral Feature Recalibration}

Finally, the learned spectral weights are used to adaptively recalibrate the input feature map via channel-wise modulation:
\begin{equation}
\mathbf{F}^{\prime}_i = \mathbf{s} \odot \mathbf{F}_i, \quad i = 1, \dots, N,
\end{equation}
where \( \odot \) denotes element-wise multiplication. This operation selectively enhances discriminative spectral channels while suppressing less informative or noisy bands, leading to more robust spectral--spatial representations.

The DSX block introduces minimal computational overhead while significantly improving the model’s ability to exploit spectral correlations inherent in hyperspectral data. By integrating dual-pooling statistics with nonlinear squeeze--expansion gating, the proposed DSX block effectively complements transformer-based attention mechanisms and plays a key role in boosting the classification performance of the DSXFormer framework.

\subsection{DSXFormer Mechanism}
The proposed DSXFormer framework is specifically designed for HSIC by integrating localized self-attention with hierarchical spectral--spatial feature modeling. At its core, DSXFormer combines a window-based transformer encoder with the proposed DSX block, enabling effective spectral recalibration and context-aware representation learning.

The encoding process begins with Layer Normalization (LN), which standardizes the input feature distributions and stabilizes the training process:
\begin{equation}
\text{LN}(x) = \frac{x - \mu}{\sqrt{\sigma^2 + \epsilon}} \gamma + \beta,
\end{equation}
where \( \mu \) and \( \sigma \) denote the mean and standard deviation of the input feature \( x \), respectively, while \( \gamma \) and \( \beta \) are learnable parameters that scale and shift the normalized feature. LN mitigates internal covariate shift and improves convergence behavior in deep transformer architectures.

Following normalization, the input feature map is partitioned into a set of non-overlapping spatial windows to facilitate localized processing. This window partitioning operation is defined as:
\begin{equation}
\mathbf{F}_{\text{windows}} = \left\{ \mathbf{F}_{i,j} \mid i = 1, \dots, m;\; j = 1, \dots, n \right\},
\end{equation}
where \( m \) and \( n \) represent the number of windows along the height and width dimensions, respectively. By constraining self-attention to local windows, DSXFormer efficiently captures fine-grained spectral--spatial patterns while maintaining favorable computational complexity. Contextual dependencies across windows are subsequently modeled through window-based multi-head self-attention with dynamic context modulation.

\subsubsection{Window-Based Dynamic Context Attention (DCA)}
The window-based DCA mechanism operates on local spatial regions to extract discriminative spectral--spatial representations while implicitly incorporating global contextual cues. Compared with global self-attention, this strategy significantly reduces computational cost and is particularly well suited for high-dimensional HSIs. Given an input feature map as:

\begin{equation}
\mathbf{F} \in \mathbb{R}^{R \times C \times d},
\end{equation}
The feature map is divided into non-overlapping windows of size \( w \times w \):
\begin{equation}
\text{windows} = \text{WindowPartition}(\mathbf{F}, w).
\end{equation}
Each window is reshaped into a sequence of tokens:
\begin{equation}
\mathbf{F}_{\text{window}} \in \mathbb{R}^{(w^2) \times d}
\end{equation}

The token sequence is linearly projected into query (\(Q\)), key (\(K\)), and value (\(V\)) matrices:
\begin{equation}
Q = \mathbf{F}_{\text{window}} W_q,\quad
K = \mathbf{F}_{\text{window}} W_k,\quad
V = \mathbf{F}_{\text{window}} W_v
\end{equation}
where \( W_q, W_k, W_v \in \mathbb{R}^{d \times d} \) are learnable projection matrices. For multi-head attention, the channel dimension is evenly divided into \( h \) heads, each with dimensionality:
\begin{equation}
d_{\text{head}} = \frac{d}{h}.
\end{equation}

The attention scores are computed using scaled dot-product attention with relative positional encoding:
\begin{equation}
A = \frac{Q K^\top}{\sqrt{d_{\text{head}}}} + B_{\text{rel}},
\end{equation}
where \( B_{\text{rel}} \) encodes relative positional information within each window, enhancing spatial awareness.

\subsubsection{Dynamic Context Scaling}
While standard self-attention relies solely on pairwise token interactions, DSXFormer introduces a Dynamic Context Scaling (DCS) mechanism to adaptively modulate attention responses based on aggregated contextual information within each window. Specifically, a context vector \( g \) is derived by averaging the raw attention scores across all tokens in a window. Let \( A_i \) denote the \( i \)-th row of the attention matrix \( A \); the context vector is defined as:
\begin{equation}
g = \frac{1}{w^2} \sum_{i=1}^{w^2} A_i.
\end{equation}
This vector summarizes the global attention tendency of the window and serves as a contextual descriptor. The raw attention scores are then adaptively scaled using the context vector:
\begin{equation}
A_{\text{scaled}} = A \odot g,
\end{equation}
where \( \odot \) denotes element-wise multiplication. This operation amplifies attention responses associated with globally informative tokens while suppressing less relevant interactions. The scaled attention scores are normalized using the softmax function:
\begin{equation}
A_{\text{softmax}} = \text{softmax}(A_{\text{scaled}}),
\end{equation}
and the final output of the attention module is obtained as:
\begin{equation}
O = A_{\text{softmax}} V.
\end{equation}

The output \( O \) is subsequently passed through a linear projection and dropout layer, followed by residual connections, forming a robust and context-aware feature representation. By combining localized attention with dynamic context scaling, DSXFormer effectively balances fine-grained discrimination and global contextual modeling for HSIC.

\subsubsection{Multi-Layer Perceptron (MLP)}

Following the window-based DCA module, a Multi-Layer Perceptron (MLP) is employed to further refine the extracted spectral--spatial representations. The MLP constitutes a critical component of the transformer encoder, enabling nonlinear feature transformation and enhancing representation capacity beyond linear attention operations. Specifically, the MLP consists of two fully connected layers separated by a Gaussian Error Linear Unit (GELU) activation function, which facilitates smooth gradient propagation and improved optimization behavior. The MLP operation is mathematically expressed as:
\begin{equation}
\text{MLP}(f) = W_2 \cdot \text{GELU}(W_1 \cdot f + b_1) + b_2,
\end{equation}
where \( W_1 \) and \( W_2 \) denote learnable weight matrices, and \( b_1 \) and \( b_2 \) represent the corresponding bias terms. Compared to conventional activation functions such as ReLU, GELU has been shown to offer superior performance in deep transformer architectures by enabling more stable training dynamics and improved feature expressiveness \cite{31hendrycks2016gaussian}.

In the context of HSIC, the MLP plays a pivotal role in capturing higher-order spectral interactions and nonlinear dependencies across bands, thereby transforming the attention-enhanced features into a more discriminative embedding space. This capability is essential for distinguishing spectrally similar classes that exhibit subtle spectral variations.

In addition to the MLP, the representational capacity of DSXFormer is further strengthened through the integration of the proposed DSX block. The DSX block adaptively recalibrates channel-wise spectral responses by modeling global spectral statistics, enabling the network to emphasize informative spectral bands while suppressing redundant or noisy features. A detailed formulation of the DSX block is provided in Section~C.

\subsubsection{Patch Extraction, Embedding, and Merging}

Patch-based operations constitute the foundation of the DSXFormer architecture, enabling efficient processing of high-dimensional hyperspectral data through structured spectral-spatial representation learning. The pipeline begins with a Patch Extraction operation, which divides the input HSI into non-overlapping spatial patches while preserving local spectral continuity.

Each extracted patch is subsequently transformed via a Patch Embedding layer, which projects the raw spectral vectors into a higher-dimensional latent space. This embedding process enhances the spectral expressiveness of each patch and converts the HSI into a tokenized representation suitable for transformer-based modeling \cite{32chen2021exploring}.

To progressively aggregate contextual information and reduce spatial redundancy, a Patch Merging operation is applied at deeper stages of the network. This operation combines embeddings from neighboring patches and projects them into a new feature space:
\begin{equation}
\begin{split}
\mathbf{x}_{\text{mg}} = \text{Concat}(&\mathbf{x}_{\text{emb}}^{(i,j)}, \mathbf{x}_{\text{emb}}^{(i+1,j)}, \\
&\mathbf{x}_{\text{emb}}^{(i,j+1)}, \mathbf{x}_{\text{emb}}^{(i+1,j+1)}) \mathbf{W},
\end{split}
\end{equation}
where \( \mathbf{x}_{\text{emb}}{(\cdot)} \) denotes the embedded patch features at the specified spatial locations, \( \text{Concat}(\cdot) \) represents channel-wise concatenation, and \( \mathbf{W} \) is a learnable projection matrix. By merging adjacent patches and applying linear projection, this operation constructs hierarchical representations that jointly encode local spectral details and broader spatial context. The patch merging strategy effectively reduces computational complexity while enhancing spectral--spatial feature abstraction, making it particularly well suited for large-scale HSIs.

\subsubsection{Prediction Head and Output Generation}

The final stage of the DSXFormer architecture is the prediction head, which maps the learned high-level representations to class probabilities for HSIC. This is achieved using a fully connected output layer followed by a softmax activation function.

Formally, the prediction process is defined as:
\begin{equation}
\text{Output}(f) = \text{softmax}(W_{\text{out}} \cdot f + b_{\text{out}}),
\end{equation}
where \( f \in \mathbb{R}^{d} \) denotes the feature vector obtained from the final transformer block, \( W_{\text{out}} \in \mathbb{R}^{d \times K} \) is the classification weight matrix, and \( b_{\text{out}} \in \mathbb{R}^{K} \) is the bias term. The softmax function is given by:
\begin{equation}
\text{softmax}(z_i) = \frac{e^{z_i}}{\sum_{j=1}^{K} e^{z_j}},
\end{equation}
which normalizes the logits to yield a valid probability distribution over the \( K \) predefined classes.

The resulting predictions form a classification map \( \mathbf{Y} \in \{1, \dots, K\}^{R \times C} \), where each pixel is assigned to the class with the highest posterior probability. When patch-level predictions are produced, upsampling techniques such as transposed convolution or interpolation can be employed to restore the output to the original spatial resolution of the input HSI.

This prediction head establishes an effective mapping between the learned spectral–spatial representations and the discrete land-cover categories, thereby completing the DSXFormer pipeline. An algorithmic overview of the proposed DSXFormer framework is presented in Algorithm~1.

\begin{algorithm}[H]
\begin{minipage}{0.9\linewidth}
\SetAlgoLined
\SetKwInOut{Input}{Input}
\SetKwInOut{Output}{Output}

\Input{HSI cube $\mathbf{H} \in \mathbb{R}^{R \times C \times L}$, spectral patch size $w$, number of spectral bands $L$, and number of attention layers}
\Output{Predicted class labels}

\BlankLine
Initialize the DSXFormer framework, drop-path regularization, DCA module, transformer layers, patch extraction, patch embedding, and patch merging modules\;

\BlankLine
\For{each spatial--spectral patch extracted from $\mathbf{H}$}{
    Extract local patches using the patch extraction layer\;
    Project the extracted patches into the latent embedding space via the patch embedding layer\;
    
    \For{each embedded patch}{
        \textbf{DSXFormer Processing:}
        \begin{enumerate}
            \item Apply the DSX block to model spectral dependencies (Eqs.~(1)--(7))\;
            \item Perform layer normalization using LN (Eq.~(8))\;
            \item Capture adaptive contextual relationships via the DCA module (Eqs.~(10)--(19))\;
            \item Transform features using the MLP block (Eq.~(20))\;
            \item Aggregate hierarchical representations using the patch merging layer (Eq.~(21))\;
            \item Generate final class predictions through the classification head (Eqs.~(22)--(23))\;
        \end{enumerate}
    }
}

\caption{Pseudocode of the proposed DSXFormer method for HSIC}
\end{minipage}
\end{algorithm}

\section{Experimental Evaluation}
In this section, we conduct a comprehensive evaluation of the proposed DSXFormer using four widely adopted benchmark HSIC datasets: Indian Pines (IP) \cite{35xu2016fusion}, Pavia University (PU) \cite{34huang2009comparative}, Salinas (SA)\footnote{http://lesun.weebly.com/hyperspectral-data-set.html}, and Kennedy Space Center (KSC) \cite{36kennedy_space_center_dataset}. We first describe the experimental setup, including dataset preprocessing procedures and evaluation protocols. Subsequently, the hyperparameter configurations employed in the proposed method are described in detailed in this article, followed by a quantitative presentation of the experimental results. A thorough comparative analysis is then performed to benchmark the proposed approach against state-of-the-art (SOTA) methods. Furthermore, we investigate the influence of key hyperparameters through ablation studies and visual analyses. This comprehensive experimental framework provides deep insights into the effectiveness, robustness, and practical applicability of the proposed method for hyperspectral image analysis.

\subsection{Datasets for Empirical Evaluation}

\begin{figure}
\centering
\subfloat[]{\includegraphics[width=0.8in,height=1.6in]{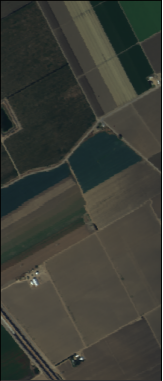}} 
\hspace{0.5mm}
\subfloat[]{\includegraphics[width=0.8in,height=1.6in]{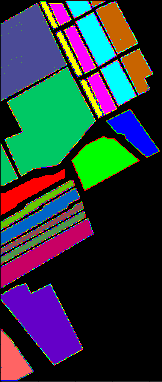}}
\hspace{0.5mm}
\subfloat[]{\includegraphics[width=1.6 in,height=0.8in]{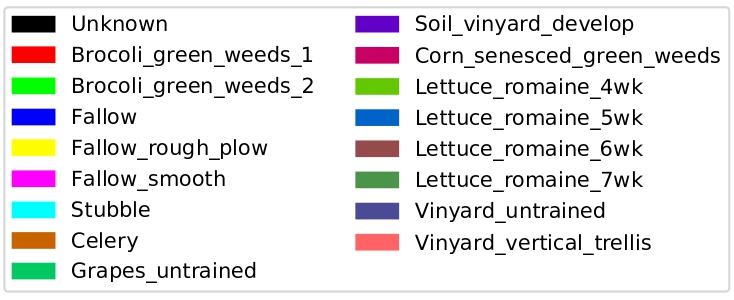}}
\hspace{0.5mm}
\caption{Illustration of the SA dataset: (a) Original image, (b) Ground truth.}
\label{fig:G1}
\end{figure}

\begin{figure}[t]
\centering
\subfloat[]{\includegraphics[width=0.8in,height=1.6in]{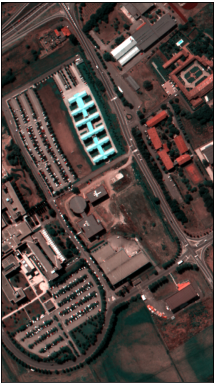}} 
\hspace{0.5mm}
\subfloat[]{\includegraphics[width=0.8in,height=1.6in]{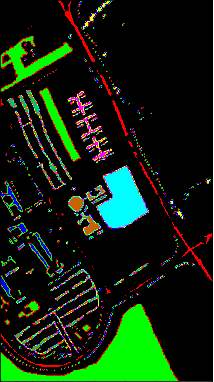}}
\hspace{0.5mm}
\subfloat[]{\includegraphics[width=1.5in,height=0.8in]{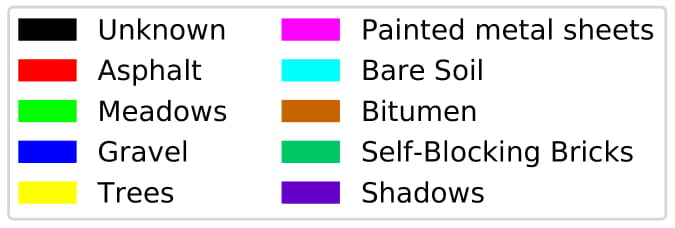}}
\hspace{0.5mm}
\caption{Illustration of the PU dataset: (a) Original image, (b) Ground truth.}
\label{fig:G2}
\end{figure}

\begin{figure}
\centering
\subfloat[]{\includegraphics[width=0.9in,height=0.9in]{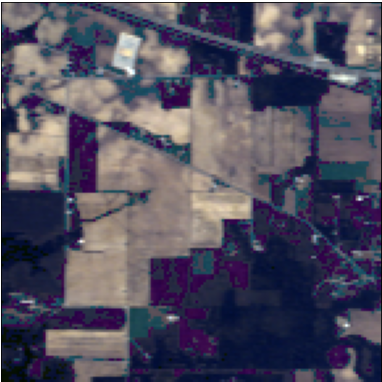}} 
\hspace{0.5mm}
\subfloat[]{\includegraphics[width=0.9in,height=0.9in]{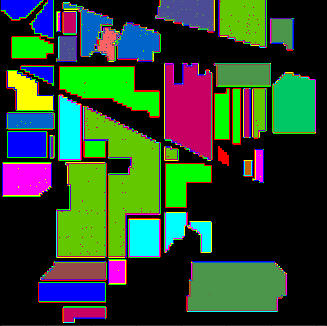}}
\hspace{0.5mm}
\subfloat[]{\includegraphics[width=1.2 in,height=0.7in]{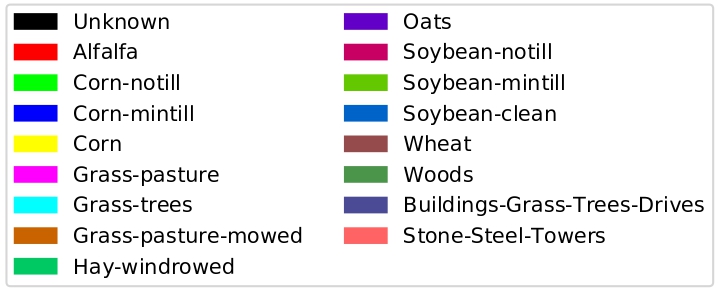}}
\hspace{0.5mm}
\caption{Illustration of the IP dataset: (a) Original image, (b) Ground truth.}
\label{fig:G3}
\end{figure}

\begin{figure}
\centering
\subfloat[]{\includegraphics[width=0.9in,height=0.9in]{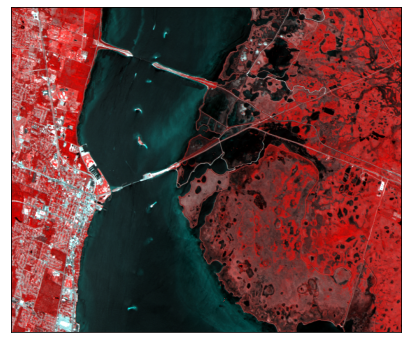}} 
\hspace{0.5mm}
\subfloat[]{\includegraphics[width=0.9in,height=0.9in]{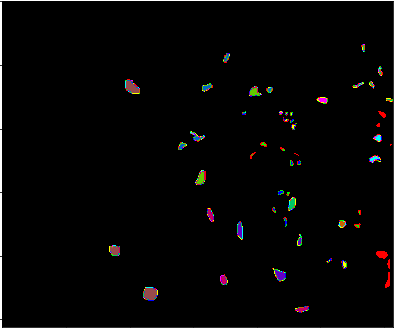}}
\hspace{0.5mm}
\subfloat[]{\includegraphics[width=1.2 in,height=0.8in]{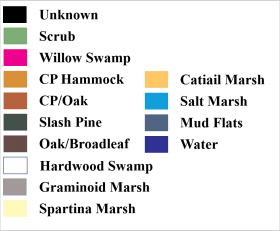}}
\hspace{0.5mm}
\caption{Illustration of the KSC dataset: (a) Original image, (b) Ground truth.}
\label{fig:G4}
\end{figure}

To comprehensively evaluate the effectiveness of the proposed DSXFormer, experiments are conducted on four widely used hyperspectral benchmark datasets, namely Salinas (SA), Pavia University (PU), Indian Pines (IP), and Kennedy Space Center (KSC). These datasets differ significantly in spatial resolution, spectral dimensionality, and scene complexity, thereby providing a rigorous and representative testbed for assessing the spectral–spatial modeling capability and generalization performance of the proposed framework. The corresponding false-color images and ground truth annotations are illustrated in Figs. \ref{fig:G1}–\ref{fig:G4}.

The SA dataset was acquired by the AVIRIS sensor over the Salinas Valley, California, with a spatial resolution of 3.7 m and 224 spectral bands spanning 360–2500 nm. It contains $512\times217$ pixels and 16 agricultural classes. The PU dataset was collected over Pavia, Italy, using the ROSIS sensor, consisting of $640\times340$ pixels at a spatial resolution of 1.3 m. After discarding noisy bands, 103 spectral channels covering 430–860 nm are retained, forming nine urban land-cover classes. 

The IP dataset, captured by AVIRIS over northwestern Indiana, comprises $145\times145$ pixels with 200 spectral bands in the range of 400–2500 nm and includes 16 crop-related classes. The KSC dataset, also acquired by AVIRIS, consists of $512\times614$ pixels with 176 effective spectral bands after noise removal and contains 13 land-cover classes representing upland and wetland environments.

The detailed dataset specifications are summarized in Table \ref{tab:table details}, and the class-wise sample distributions with corresponding training and testing splits are reported in Tables \ref{tab:table SA}–\ref{tab:table KSC}. These datasets collectively enable a comprehensive and fair evaluation of DSXFormer, particularly in terms of its ability to capture discriminative spectral dependencies while maintaining robust spatial feature representation.

\begin{table}
\centering
\captionsetup{justification=centering,singlelinecheck=off}
\caption{DETAILED DESCRIPTION OF EACH DATASET USED DURING EXPERIMENT}
\label{tab:table details}
\resizebox{\columnwidth}{!}{%
\begin{tabular}{ccccc}
\hline 
Name & Spatial Dimension & Spectral Bands & Wavelength Range & Classes \\ \hline
IP   & 145x145           & 224            & 400nm - 2500nm   & 16      \\
PU   & 610x340           & 103            & 430nm - 860nm    & 9       \\
SA   & 512x217           & 224            & 360nm - 2500nm   & 16      \\
KSC  & 610x340           & 224            & 400nm - 2500nm   & 13      \\ \hline
\end{tabular}%
}
\end{table}

\begin{table}
\centering
\captionsetup{justification=centering,singlelinecheck=off}
\caption{AMOUNTS OF TRAINING AND TEST DATA OF THE SA DATASET}
\label{tab:table SA}
\resizebox{\columnwidth}{!}{%
\begin{tabular}{cccc}
\hline
 Class   No. & Class Name                & Training & Test  \\ \hline
1           & Brocoli\_green\_weeds\_1     & 201      & 1808  \\
2           & Brocoli\_green\_weeds\_2     & 373      & 3353  \\
3           & Fallow                    & 198      & 1778  \\
4           & Fallow\_rough
\_plow         & 140      & 1254  \\
5           & Fallow\_smooth             & 268      & 2410  \\
6           & Stubble                   & 396      & 3563  \\
7           & Celery                    & 358      & 3221  \\
8           & Grapes\_untrained          & 1128     & 10143 \\
9           & Soil\_vinyard\_develop      & 621      & 5582  \\
10          & Corn\_senesced\_green\_weeds & 328      & 2950  \\
11          & Lettuce\_romaine\_4wk       & 107      & 961   \\
12          & Lettuce\_romaine\_5wk       & 193      & 1734  \\
13          & Lettuce\_romaine\_6wk       & 92       & 824   \\
14          & Lettuce\_romaine\_7wk       & 107      & 963   \\
15          & Vinyard\_untrained         & 727      & 6,541 \\
16          & Vinyard\_vertical\_trellis  & 181      & 1626  \\ \hline  
         & Total                     & 5418     & 48711 \\ \hline
\end{tabular}%
}
\end{table}

\begin{table}
\centering
\captionsetup{justification=centering,singlelinecheck=off}
\caption{AMOUNTS OF TRAINING AND TEST DATA OF THE PU DATASET}
\label{tab:table PU}
\resizebox{\columnwidth}{!}{%
\begin{tabular}{cccc}
\hline 
 Class   No. & Class Name     & Training & Test  \\ \hline 
1           & Asphalt        & 664      & 5967  \\
2           & Medows         & 1865     & 16784 \\
3           & Gravel         & 210      & 1889  \\
4           & Tress          & 307      & 2757  \\
5           & Metal   sheets & 135      & 1210  \\
6           & Bare soil      & 503      & 4526  \\
7           & Bitumen        & 133      & 1197  \\
8           & Bricks         & 369      & 3313  \\
9           & Shadows        & 664      & 5967  \\ \hline  
            & Total          & 4850     & 43610 \\ \hline 
\end{tabular}%
}
\end{table}

\begin{table}
\centering
\captionsetup{justification=centering,singlelinecheck=off}
\caption{AMOUNTS OF TRAINING AND TEST DATA OF THE IP SCENE DATASET}
\label{tab:table IP}
\resizebox{\columnwidth}{!}{%
\begin{tabular}{cccc}
\hline
 Class   No. & Class Name            & Training & Test \\ \hline
1           & Alfalfa               & 5        & 41   \\
2           & Corn-notill           & 143      & 1285 \\
3           & Corn-min              & 83       & 747  \\
4           & Corn                  & 24       & 213  \\
5           & Grass-pasture         & 49       & 434  \\
6           & Grass-trees           & 73       & 657  \\
7           & Grass-pasture-mowed   & 3        & 25   \\
8           & Hay-windrowed         & 48       & 430  \\
9           & Oats                  & 2        & 18   \\
10          & Soybean-notill        & 98       & 874  \\
11          & Soybean-mintill       & 246      & 2209 \\
12          & Soybean-clean         & 60       & 533  \\
13          & Wheat                 & 21       & 184  \\
14          & Woods                 & 127      & 1138 \\
15          & Buildings-grass-trees & 39       & 347  \\
16          & Stone-steel-towers    & 10       & 83   \\ \hline  
           & Total                 & 1031     & 9218 \\ \hline
\end{tabular}%
}
\end{table}

\begin{table}
\centering
\captionsetup{justification=centering,singlelinecheck=off}
\caption{AMOUNTS OF TRAINING AND TEST DATA OF KSC SCENE DATASET}
\label{tab:table KSC}
\resizebox{\columnwidth}{!}{%
\begin{tabular}{cccc}
\hline
 Class   No. & Class Name        & Training & Test \\ \hline
1           & Scrub             & 77       & 684   \\
2           & Willom swamp    & 25       & 218  \\
3           & CP hammock      & 26        & 230   \\
4           & Slash pine      & 26       & 226  \\
5           & Oak/broadleaf     & 17      & 144 \\
6           & Hardwood          & 23      & 206 \\
7           & Swap              & 11        & 94   \\
8           & Graminoid marsh & 44       & 387  \\
9           & Spartina marsh  & 52       & 468  \\
10          & Cattail marsh   & 41       & 363  \\
11          & Salt marsh      & 42       & 377  \\
12          & Mud flats         & 51       & 452  \\
13          & Water             & 93        & 834   \\ \hline  
           & Total             & 528     & 4683 \\ \hline
\end{tabular}%
}
\end{table}


\begin{table*}
\centering
\captionsetup{justification=centering,singlelinecheck=off} 
\caption{Classification accuracies of our proposed DSXFormer model on the SA dataset, evaluated in terms of OA, AA, and K, compared with SOTA methods.}
\label{tab:table 6-sa}
\resizebox{\linewidth}{!}{%
\begin{tabular}{cccccccccccccc}
\toprule
\textbf{No.} & \textbf{Class Names} & \textbf{SVM-RBF} & \textbf{CCF-200} & \textbf{2D CNN} \cite{37ge2020hyperspectral} & \textbf{GCNN}\cite{38hong2020graph} & \textbf{FADCNN}\cite{39yu2021feedback} & \textbf{NL-GCNN}\cite{40mou2020nonlocal} & \textbf{VIT}\cite{46ayas2022spectralswin} & \textbf{WaveFormer}\cite{50waveformer}& \textbf{PMCN}\cite{51PMCN}& \textbf{PyFormer}\cite{48pyramid}& \textbf{SwinT} \cite{47liu2023spectral}& \textbf{DSXFormer}\\ \midrule
1  & Brocoli\_green\_weeds\_1     & 98.98 & 99.49 & 71.57 & 99.59 & 88.26 & 99.69 & 99.93 & 99.72& 99.82&\textbf{100}&\textbf{100}& \textbf{100}\\
2  & Brocoli\_green\_weeds\_2     & 99.67 & 99.95 & 99.86 & 98.07 & 98.04 & 99.21 & 99.69 & 99.91& 99.75&99.93&\textbf{100}& \textbf{100}\\
3  & Fallow                       & 98.70 & 99.43 & 88.89 & 91.95 & 98.04 & 99.79 & 99.78 & 99.72& \textbf{100}&99.87&\textbf{100}& \textbf{100}\\
4  & Fallow\_rough\_plow          & 97.77 & 99.33 & 98.14 & 97.84 & 97.13 & 98.29 & 85.35 & 97.40& 98.22&98.21&98.78& \textbf{99.90}\\
5  & Fallow\_smooth               & 98.33 & 98.82 & 98.17 & 98.06 & 99.13 & 99.28 & 88.67 & 98.40& 99.29&99.39&\textbf{99.95}& 99.63\\
6  & Stubble                      & 99.72 & 99.80 & \textbf{100}   & 99.00 & 99.10 & 99.80 & 99.78 & 99.94& 99.79&99.97&99.64& \textbf{100}\\
7  & Celery                       & 99.46 & 99.66 & 97.00 & 99.29 & 99.21 & 99.04 & \textbf{100} & 99.97& 99.97&\textbf{100}&\textbf{100}& 99.76\\
8  & Grapes\_untrained            & 70.37 & 67.56 & 70.79 & 82.25 & 75.72 & 79.11 & 86.85 & 99.08& 99.49&99.61&99.95& \textbf{100}\\
9  & Soil\_vinyard\_develop       & 98.59 & 99.19 & 99.45 & 97.11 & \textbf{100}   & 97.74 & \textbf{100} & 99.95& 99.96&\textbf{100}&\textbf{100}& \textbf{100}\\
10 & Corn\_senesced\_green\_weeds & 93.74 & 93.80 & 96.19 & 91.60 & 79.00 & 95.01 & 99.49 & 99.80& 99.86&99.85&\textbf{100}& \textbf{100}\\
11 & Lettuce\_romaine\_4wk        & 94.70 & 95.87 & 96.37 & 90.77 & 95.3  & 94.60 & 99.03 & \textbf{99.90}& 99.67&99.77&99.73& 99.73\\
12 & Lettuce\_romaine\_5wk        & 99.89 & 99.95 & \textbf{100}   & \textbf{100}   & 98.34 & \textbf{100}   & 91.49 & 99.83& 99.57&99.55&\textbf{100}& 99.93\\
13 & Lettuce\_romaine\_6wk        & 97.81 & 98.15 & \textbf{100}   & 98.96 & 95.12 & 98.96 & 88.45 & 98.57& 99.87&99.59&\textbf{100}& \textbf{100}\\
14 & Lettuce\_romaine\_7wk        & 97.35 & 96.86 & 98.33 & 97.35 & 98.86 & 99.41 & 82.56 & 99.90& 99.89&\textbf{100}&98.94& \textbf{100}\\
15 & Vinyard\_untrained           & 71.53 & 80.77 & 91.22 & 70.44 & 99.23 & 84.26 & 86.57 & 98.77& 99.11&99.66&99.88& \textbf{99.96}\\
16 & Vinyard\_vertical\_trellis   & 98.18 & 98.18 & 93.00 & 97.10 & 83.53 & 98.01 & \textbf{100} & 99.63& 99.93&\textbf{100}&\textbf{100}& \textbf{100}\\ \hline 
OA & -                            & 88.82 & 89.72 & 90.25 & 90.37 & 90.58 & 92.28 & 93.52 & 99.40& 99.61&99.75&99.88& \textbf{99.95}\\  
AA & -                            & 94.67 & 95.43 & 93.69 & 94.34 & 94.56 & 96.39 & 94.48 & 99.29& 99.59&99.68&99.83& \textbf{99.93}\\  
K  & -                            & 87.57 & 88.58 & 89.18 & 89.28 & 90.25 & 89.64 & 92.78 & 99.33& 99.57&99.72&99.87& \textbf{99.95}\\ \bottomrule
\end{tabular}%
}
\end{table*}

\begin{table*}
\centering
\captionsetup{justification=centering,singlelinecheck=off} 
\caption{Classification accuracies of our proposed DSXFormer model on the IP dataset, evaluated in terms of OA, AA, and K, compared with SOTA methods.}
\label{tab:table7-IP}
\resizebox{\linewidth}{!}{%
\begin{tabular}{cccccccccccccc}
\toprule
\textbf{No.} & \textbf{Class Names} & \textbf{SVM-RBF} & \textbf{CCF-200} & \textbf{2D CNN} \cite{37ge2020hyperspectral} & \textbf{GCNN}\cite{38hong2020graph} & \textbf{FADCNN}\cite{39yu2021feedback} & \textbf{NL-GCNN}\cite{40mou2020nonlocal} & \textbf{VIT}\cite{46ayas2022spectralswin} & \textbf{WaveFormer}\cite{50waveformer}& \textbf{PMCN}\cite{51PMCN}& \textbf{PyFormer}\cite{48pyramid}& \textbf{SwinT} \cite{47liu2023spectral}& \textbf{DSXFormer}\\ \midrule
1  & Alfalfa                      & 71.39 & 76.37 & 54.77 & 76.66 & 88.26 & 83.09 & \textbf{100}&  97.06&93.75&93.75&83.78& \textbf{100}\\
2  & Corn-notill                  & 71.05 & 77.93 & 96.94 & 86.10 & \textbf{98.04}& 89.03 & 67.51 &  91.88&92.70&96.00&94.97& 98.50\\
3  & Corn-mintill                 & 86.96 & 94.57 & 99.46 & \textbf{100}& 97.04 & \textbf{100}& 74.47 &  97.91&98.45&98.11&90.76& 98.97\\
4  & Corn                         & 91.72 & 94.41 & 96.87 & 93.06 & 96.03 & 93.51 & \textbf{100}&  93.82&83.13&98.19&95.65& \textbf{100}\\
5  & Grass-pasture                & 85.80 & 91.39 & 94.12 & 92.06 & \textbf{99.34}& 94.12 & 93.43 &  93.92&96.75&95.27&96.60& 95.56\\
6  & Grass-trees                  & 93.85 & 97.04 & 96.81 & 96.81 & 99.48& 98.18 & 86.58 &  95.26&\textbf{99.61}&98.24&92.15& 99.02  
\\
7  & Grass-pasture-mowed          & 75.38 & 90.96 & 91.29 & 88.24 & 75.72 & 88.24 & \textbf{100}&  47.62&70.00&70.00&90.91& \textbf{100.00}\\
8  & Hay-windrowed                & 59.88 & 69.48 & 93.05 & 76.80 & \bf{100}   & 78.78 & 97.95 &  99.16&\textbf{100}&\textbf{100}&99.70& \textbf{100}\\
9  & Oats                         & 76.24 & 89.01& 87.59 & 80.85 & 78.00 & 86.70 & 0.00 &  40.00&42.86&50.00&66.67& 64.29\\
10 & Soybean-notill               & 96.91 & 98.77 & \textbf{100}& 99.38 & 96.50 & 99.38 & 86.12 &  91.36&94.41&92.65&94.79& 98.82  
\\
11 & Soybean-mintill              & 79.58 & 93.73 & 68.57 & 93.89 & 98.49 & 94.94 & 90.68 &  98.53&98.31&97.91&93.59& \textbf{99.42}\\
12 & Soybean-clean                & 74.84 & 74.55 & 88.48 & 93.64 & 95.45 & 97.27& 80.79 &  95.73&91.08&95.90&92.29& \textbf{99.04}\\
13 & Wheat                        & 97.78 & \textbf{100}& \textbf{100}& \textbf{100}& 99.02 & \textbf{100}& 87.12 &  86.36&\textbf{100}&93.01&95.14& 97.90  
\\
14 & Woods                        & 79.49 & 97.44 & 82.05 & 92.31 & 99.48 & 97.44 & 96.67 &  \textbf{100}&99.44&99.66&98.88& \textbf{100}\\
15 & Buildings-Grass-Trees-Drives & \textbf{100}& 90.91 & \textbf{100}& \textbf{100}& 99.74 & \textbf{100}& 98.28 &  95.16&98.52&96.30&98.48& \textbf{100}\\
16 & Stone-Steel-Towers           & \textbf{100}& \textbf{100}& \textbf{100}& \textbf{100}& 82.80 & \textbf{100}& 66.67 &  62.86&95.38&70.77&78.57& 89.23  
\\ \hline  
OA & -                            & 74.24 & 82.87 & 90.25 & 85.43 & 97.80 & 87.92 & 85.44 &  95.41&96.45&96.67&94.61& \textbf{98.91}\\  
AA & -                            & 83.80 & 89.78 & 93.69 & 91.87 & 93.96 & 93.79 & 66.30 &  86.66&90.89&90.36&85.42& \textbf{96.30}\\  
K  & -                            & 70.93 & 80.59 & 89.18 & 83.42 & 86.10 & 86.25 & 83.32 &  94.76&95.94&96.20&93.84& \textbf{98.76}\\ \bottomrule
\end{tabular}%
}
\end{table*}

\begin{table*}
\centering
\captionsetup{justification=centering,singlelinecheck=off} 
\caption{Classification accuracies of our proposed DSXFormer model on the PU dataset, evaluated in terms of OA, AA, and K, compared with SOTA methods.}
\label{tab:table8-PU}
\resizebox{\linewidth}{!}{%
\begin{tabular}{cccccccccccclc}
\toprule
\textbf{No.} & \textbf{Class Names} & \textbf{SVM-RBF} & \textbf{CCF-200} & \textbf{2D CNN} \cite{37ge2020hyperspectral} & \textbf{GCNN}\cite{38hong2020graph} & \textbf{FADCNN}\cite{39yu2021feedback} & \textbf{NL-GCNN}\cite{40mou2020nonlocal} & \textbf{VIT}\cite{46ayas2022spectralswin} & \textbf{WaveFormer}\cite{50waveformer}& \textbf{PMCN}\cite{51PMCN}& \textbf{PyFormer}\cite{48pyramid}& \textbf{SwinT} \cite{47liu2023spectral}& \textbf{DSXFormer}\\ \midrule
1  & Asphalt                & 82.37 & 86.59 & 83.85 & 78.89 & 99.70& 86.80 & 87.64&   99.03&99.40&99.26&99.36 & \textbf{100}\\
2  & Meadows                & 67.87 & 72.33 & 96.09 & 90.50 & 99.75 & 88.74 & 98.70&   \textbf{100}&\textbf{100}&\textbf{100}&99.81 & 99.99  
\\
3  & Gravel                 & 69.18 & 71.75 & 81.47 & 71.70 & 94.31 & 70.84 & 86.50&   96.95&96.32&98.03&98.85& \textbf{99.73}\\
4  & Trees                  & 98.37 & 99.09 & 96.12 & 98.76 & \textbf{99.10}& 98.43 & 80.73&   97.61&97.78&97.43&98.80 & 98.32  
\\
5  & Painted   metal sheets & 99.41 & 99.78 & 98.74 & 99.93& 99.82 & 99.85 & 85.87&   \textbf{100}&\textbf{100}&\textbf{100}&97.21 & \textbf{100}\\
6  & Bare Soil              & 93.64 & 97.26 & 49.79 & 79.08 & 99.92 & 94.37 & 99.06&   99.71&99.79&99.76&99.94& \textbf{100}\\
7  & Bitumen                & 91.20 & 91.88 & 79.32 & 71.20 & 96.97 & 86.24 & 93.28&   \textbf{100}&\textbf{100}&\textbf{100}&98.83 & 99.89\\
8  & Self-Blocking   Bricks & 92.59 & 94.92 & 88.89 & 92.83 & 97.90 & 96.74 & 84.89&   93.52&94.31&94.39&98.73& \textbf{100}  
\\
9  & Shadows                & 96.94 & 98.73 & 94.19 & 97.47 & 98.97 & 95.78 & 88.31&   98.17&\textbf{99.58}&98.31&98.85 & 99.40  
\\ 
\hline  
OA & -                      & 78.89 & 83.36 & 86.93 & 87.08 & 98.87& 90.04 & 93.31&   98.90&99.04&99.06&99.41 & \textbf{99.85}\\  
AA & -                      & 87.95 & 90.26 & 85.38 & 86.71 & 98.05& 90.87 & 85.96&   98.33&98.57&98.57&98.45 & \textbf{99.70}\\  
K  & -                      & 74.91 & 79.05 & 82.42 & 83.07 & 95.25 & 87.06 & 91.10&   98.54&98.73&98.75&99.22 & \textbf{99.80}\\ \bottomrule
\end{tabular}%
}
\end{table*}

\begin{table*}
\centering
\caption{Classification accuracies of our proposed DSXFormer model on the KSC dataset, evaluated in terms of OA, AA, and K, compared with SOTA methods.}
\label{tab:table 9-ksc}
\resizebox{\linewidth}{!}{%
\begin{tabular}{@{}cccccccclccclc@{}}
\toprule
\textbf{Class No.} &  \textbf{Class Names}&\textbf{SVM-RBF} & \textbf{CCF-200} & \textbf{2D CNN}\cite{37ge2020hyperspectral} & \textbf{GCNN}\cite{38hong2020graph} & \textbf{FADCNN}\cite{39yu2021feedback} & \textbf{NL-GCNN}\cite{40mou2020nonlocal} & \textbf{VIT}\cite{46ayas2022spectralswin} & \textbf{WaveFormer}\cite{50waveformer}& \textbf{PMCN}\cite{51PMCN}& \textbf{PyFormer}\cite{48pyramid}& \textbf{SwinT}\cite{47liu2023spectral}& \textbf{DSXFormer}\\ \midrule
1                  &  Scrub             
&95.10            & 96.23            & \textbf{100}    & 90.14         & 97.33           & 72.34             & 96.17 &  98.98&99.34&\textbf{100}&92.18& 98.69\\
2                  &  Willom swamp    
&62.14            & \textbf{97.21}& 85.67           & 91.31         & 61.25           & 93.41             & 86.47 &  90.87&87.11&92.35&96.00& 98.24  
\\
3                  &  CP hammock      
&89.12            & 71.31            & 73.59           & 73.24         & 66.37           & 81.44             & 92.61 &  94.35&\textbf{96.10}&92.74&82.47& 95.53  
\\
4                  &  Slash pine      
&74.19            & 66.15            & 73.45           & 71.35         & 68.91           & 73.87             & 81.22 &  94.71&95.54&\textbf{97.73}&81.87& \textbf{97.73}  
\\
5                  &  Oak/broadleaf     
&69.23            & 00.69            & 93.19           & 81.42         & 65.16           & 90.65             & \textbf{99.00}&  84.83&79.84&83.19&95.60& 86.73  
\\
6                  &  Hardwood          
&53.21            & 00.24            & 98.51           & 25.23         & 83.89           & 98.22             & \textbf{99.36} &  84.95&88.52&86.88&98.65& 91.87  
\\
7                  &  Swap              
&61.25            & 00.00            & 99.51           & 88.92         & 81.67           & \textbf{99.42}& 89.77 &  91.49&94.05&91.89&93.24& \textbf{100}  
\\
8                  &  Graminoid marsh 
&97.29& 78.42            & 84.33           & 79.33         & 64.88           & 71.29             & 80.67 &  93.04&\textbf{97.39}&97.02&79.94& \textbf{100}\\
9                  &  Spartina marsh  
&98.35            & 99.34            & 97.28           & 92.35         & \textbf{100}    & \textbf{100}      & 85.88 &  99.15&97.60&99.18&91.99& 99.73\\
10                 &  Cattail marsh   
&50.33            & \textbf{100}     & 99.12           & 59.15         & 89.37           & 92.66             & 99.30 &  92.58&96.90&99.29&95.19& \textbf{100}   
\\
11                 &  Salt marsh      
&00.00            & \textbf{100}     & 98.24           & 70.37         & 97.68           & 90.43             & 98.23 &  \textbf{100}&99.70&\textbf{100}&99.66& \textbf{100}   
\\
12                 &  Mud flats         
&76.38            & 89.89            & 12.43           & 83.98         & 94.27           & 99.12             & 95.83 &  96.69&97.01&98.30&96.58& \textbf{99.15}\\
13                 &  Water             
&98.43            & 82.91            & 99.28           & \textbf{100}  & \textbf{100}    & \textbf{100}      & 99.46 &  \textbf{100}&99.87&99.85&99.39& \textbf{100}\\ \midrule 
OA                 &  &78.53            & 83.81            & 84.72           & 85.56         & 87.84           & 89.57             & 93.31 &  96.03&96.62&97.34&93.12& \textbf{98.52}\\  
AA                 &  &72.26            & 69.86            & 84.31           & 78.92         & 85.26           & 90.69             & 90.82 &  93.97&94.54&95.26&90.56& \textbf{97.51}\\  
K                  &  &76.04            & 81.83            & 83.02           & 83.82         & 86.51           & 88.44             & 92.54 &  95.58&96.23&97.04&92.33& \textbf{98.35}\\ \bottomrule
\end{tabular}%
}
\end{table*}


\begin{figure*}
\centering
\label{fig.4}

\subfloat[]{\includegraphics[width=0.8in,height=1.6in]{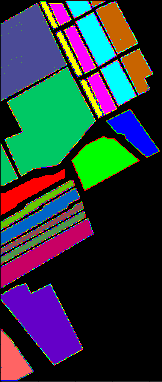}} 
\hspace{0.2mm}
\subfloat[]{\includegraphics[width=0.8in,height=1.6in]{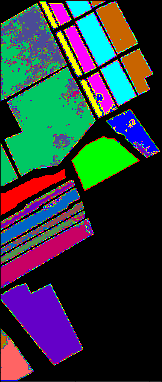}}
\hspace{0.2mm}
\subfloat[]{\includegraphics[width=0.8in,height=1.6in]{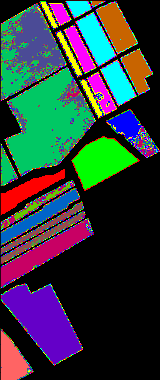}}
\hspace{0.2mm}
\subfloat[]{\includegraphics[width=0.8in,height=1.6in]{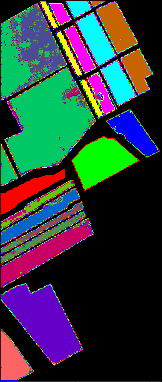}}
\hspace{0.2mm}
\subfloat[]{\includegraphics[width=0.8in,height=1.6in]{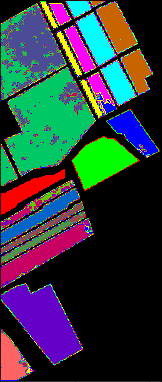}}
\hspace{0.2mm}
\subfloat[]{\includegraphics[width=0.8in,height=1.6in]{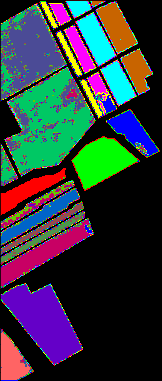}}
\hspace{0.2mm}
\subfloat[]{\includegraphics[width=0.8in,height=1.6in]{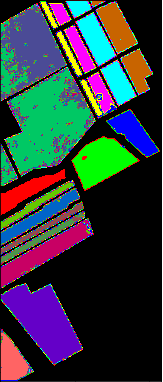}}
\hspace{0.2mm}
\subfloat[]{\includegraphics[width=0.8in,height=1.6in]{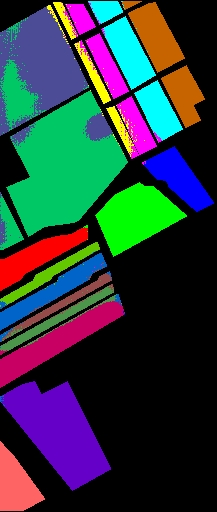}}
\hspace{0.2mm}
\subfloat[]{\includegraphics[width=0.8in,height=1.6in]{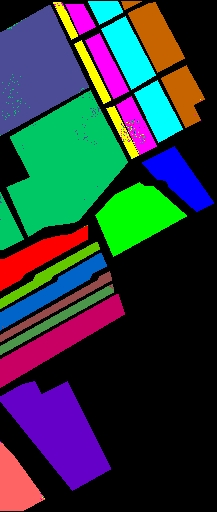}}
\hspace{0.2mm}
\subfloat[]{\includegraphics[width=0.8in,height=1.6in]{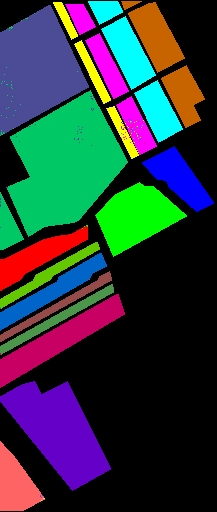}}
\hspace{0.2mm}
\subfloat[]{\includegraphics[width=0.8in,height=1.6in]{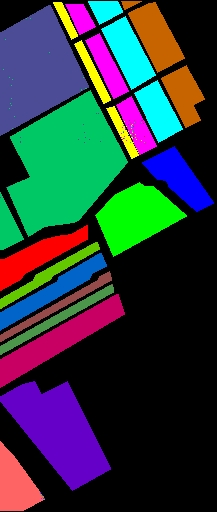}}
\hspace{0.2mm}
\subfloat[]{\includegraphics[width=0.8in,height=1.6in]{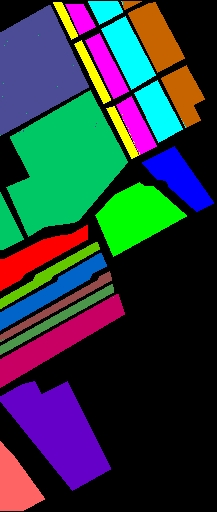}}
\hspace{0.2mm}
\subfloat[]{\includegraphics[width=0.8in,height=1.6in]{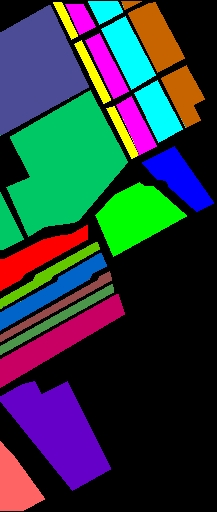}}
\caption{Shows the classification maps of various methods for the SA data set. (From Left to Right) (a) groundtruth, (b) SVM-RBF, (c) CCF-200, (d) 2-D CNN\cite{37ge2020hyperspectral}, (e) GCNN\cite{38hong2020graph}, (f) FADCNN\cite{39yu2021feedback}, (g) NL-GCNN\cite{40mou2020nonlocal}, (h) ViT\cite{46ayas2022spectralswin}, (i) WaveFormer\cite{50waveformer}, (j) PMCN\cite{51PMCN}, (k) PyFormer\cite{48pyramid}, (l) SwinT\cite{47liu2023spectral}, and (m) (Our) DSXFormer.}
\label{fig: 4}
\end{figure*}

\begin{figure*}
\centering
\subfloat[]{\includegraphics[width=0.9in,height=0.9in]{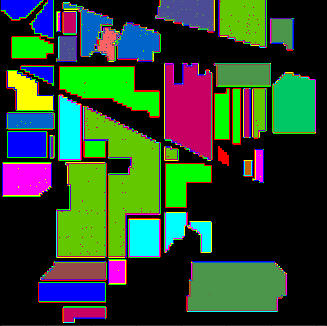}}
\hspace{0.2mm}
\subfloat[]{\includegraphics[width=0.9in,height=0.9in]{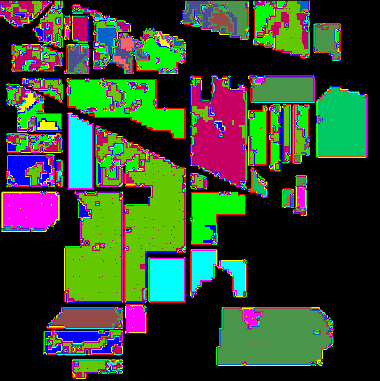}}
\hspace{0.2mm}
\subfloat[]{\includegraphics[width=0.9in,height=0.9in]{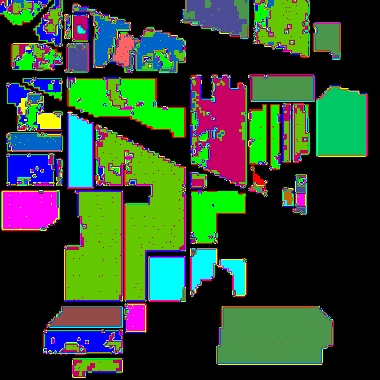}}
\hspace{0.2mm}
\subfloat[]{\includegraphics[width=0.9in,height=0.9in]{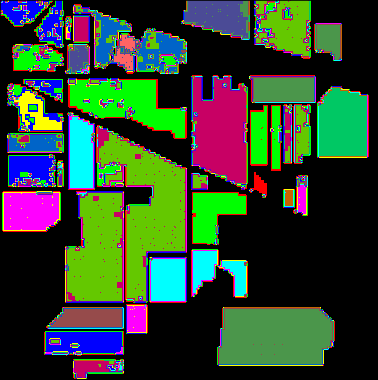}}
\hspace{0.2mm}
\subfloat[]{\includegraphics[width=0.9in,height=0.9in]{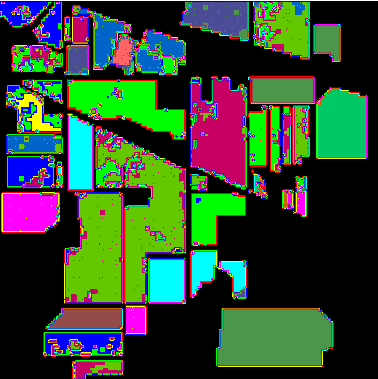}}
\hspace{0.2mm}
\subfloat[]{\includegraphics[width=0.9in,height=0.9in]{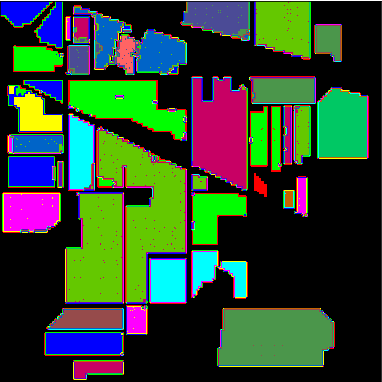}}
\hspace{0.2mm}
\subfloat[]{\includegraphics[width=0.9in,height=0.9in]{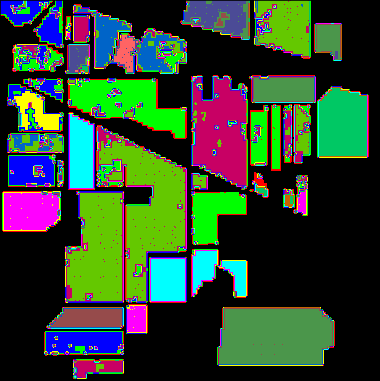}}
\hspace{0.2mm}
\subfloat[]{\includegraphics[width=0.9in,height=0.9in]{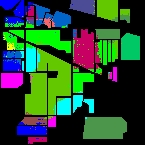}}
\hspace{0.2mm}
\subfloat[]{\includegraphics[width=0.9in,height=0.9in]{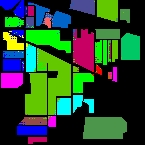}}
\hspace{0.2mm}
\subfloat[]{\includegraphics[width=0.9in,height=0.9in]{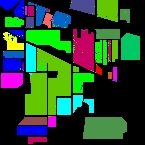}}
\hspace{0.2mm}
\subfloat[]{\includegraphics[width=0.9in,height=0.9in]{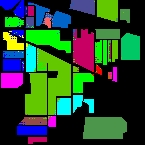}}
\hspace{0.2mm}
\subfloat[]{\includegraphics[width=0.9in,height=0.9in]{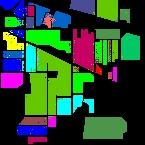}}
\hspace{0.2mm}
\subfloat[]{\includegraphics[width=0.9in,height=0.9in]{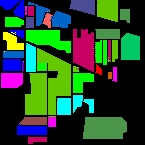}}
\caption{Shows the classification maps of various methods for the IP data set. (From Left to Right and from top to bottom) (a) ground-truth, (b) SVM-RBF, (c) CCF-200, (d) 2-D CNN\cite{37ge2020hyperspectral}, (e) GCNN\cite{38hong2020graph}, (f) FADCNN\cite{39yu2021feedback}, (g) NL-GCNN\cite{40mou2020nonlocal}, (h) ViT\cite{46ayas2022spectralswin}, (i) WaveFormer\cite{50waveformer}, (j) PMCN\cite{51PMCN}, (k) PyFormer\cite{48pyramid}, (l) SwinT\cite{47liu2023spectral}, and (m) (Our) DSXFormer.}
\label{fig: 5}
\end{figure*}

\begin{figure*}
\centering

\subfloat[]{\includegraphics[width=0.8in,height=1.6in]{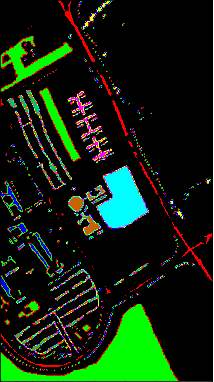}}
\hspace{0.2mm}
\subfloat[]{\includegraphics[width=0.8in,height=1.6in]{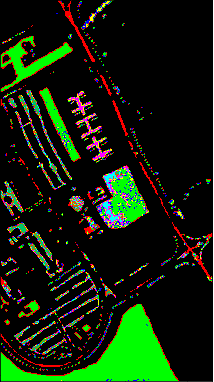}}
\hspace{0.2mm}
\subfloat[]{\includegraphics[width=0.8in,height=1.6in]{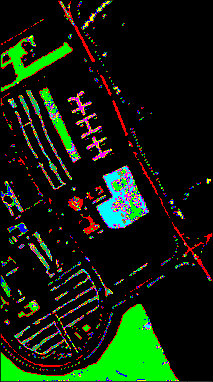}}
\hspace{0.2mm}
\subfloat[]{\includegraphics[width=0.8in,height=1.6in]{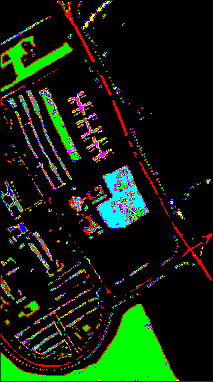}}
\hspace{0.2mm}
\subfloat[]{\includegraphics[width=0.8in,height=1.6in]{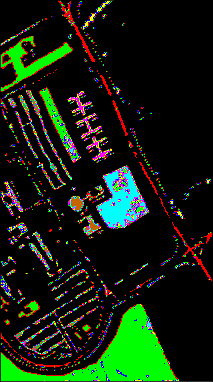}}
\hspace{0.2mm}
\subfloat[]{\includegraphics[width=0.8in,height=1.6in]{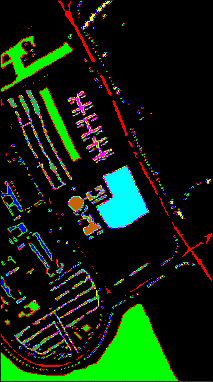}}
\hspace{0.2mm}
\subfloat[]{\includegraphics[width=0.8in,height=1.6in]{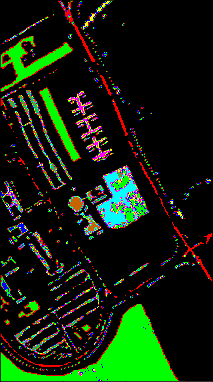}}
\hspace{0.2mm}
\subfloat[]
{\includegraphics[width=0.8in,height=1.6in]{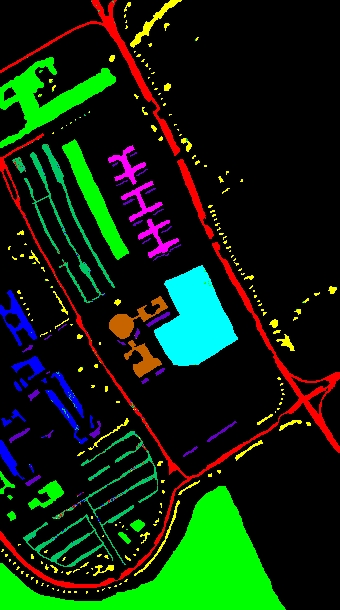}}
\hspace{0.2mm}
\subfloat[]
{\includegraphics[width=0.8in,height=1.6in]{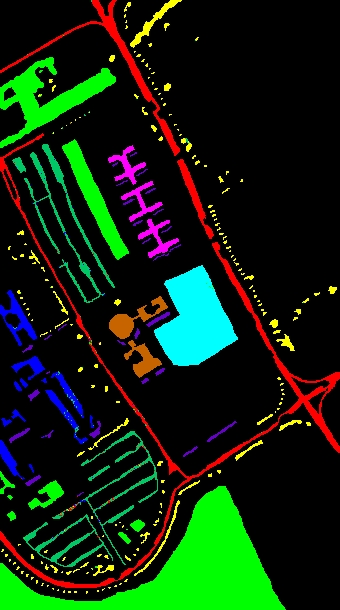}}
\hspace{0.2mm}
\subfloat[]
{\includegraphics[width=0.8in,height=1.6in]{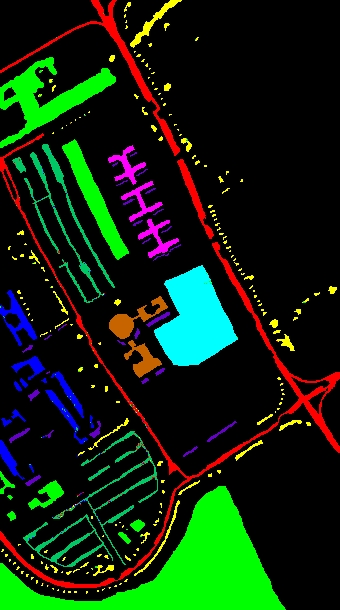}}
\hspace{0.2mm}
\subfloat[]
{\includegraphics[width=0.8in,height=1.6in]{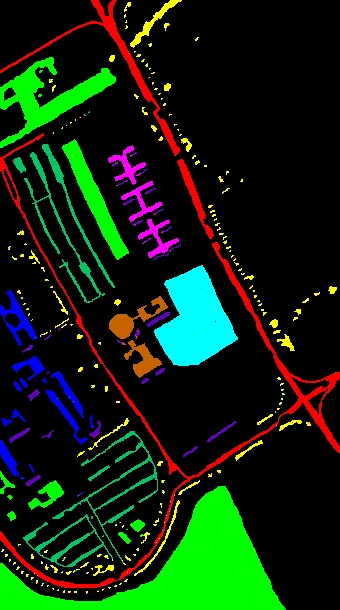}}
\hspace{0.2mm}
\subfloat[]
{\includegraphics[width=0.8in,height=1.6in]{Figures/Results/pu/30_predictions_pu_swint.jpg}}
\hspace{0.2mm}
\subfloat[]{\includegraphics[width=0.8in,height=1.6in]{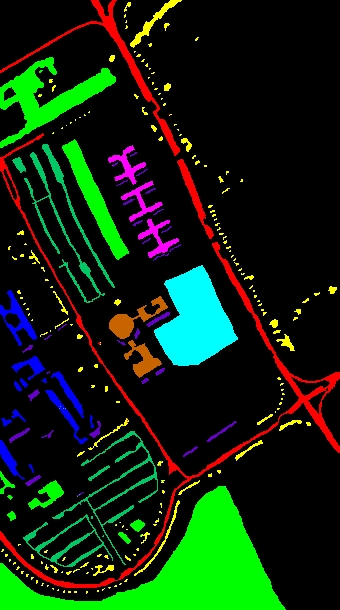}}
\caption{Shows the classification maps of various methods for the PU data set. (From Left to Right) (a) ground truth, (b) SVM-RBF, (c) CCF-200, (d) 2-D CNN\cite{37ge2020hyperspectral}, (e) GCNN\cite{38hong2020graph}, (f) FADCNN\cite{39yu2021feedback}, (g) NL-GCNN\cite{40mou2020nonlocal}, (h) ViT\cite{46ayas2022spectralswin}, (i) WaveFormer\cite{50waveformer}, (j) PMCN\cite{51PMCN}, (k) PyFormer\cite{48pyramid}, (l) SwinT\cite{47liu2023spectral}, and (m) (Our) DSXFormer.}
\label{fig: 7}
\end{figure*}

\begin{figure*}
\centering
\subfloat[]
{\includegraphics[width=0.9in,height=0.9in]{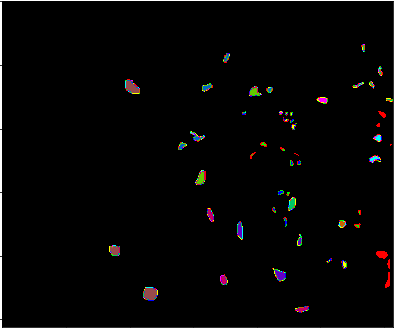}} 
\hspace{0.2mm}
\subfloat[]
{\includegraphics[width=0.9in,height=0.9in]{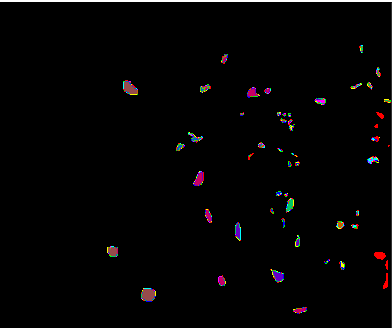}}
\hspace{0.2mm}
\subfloat[]
{\includegraphics[width=0.9in,height=0.9in]{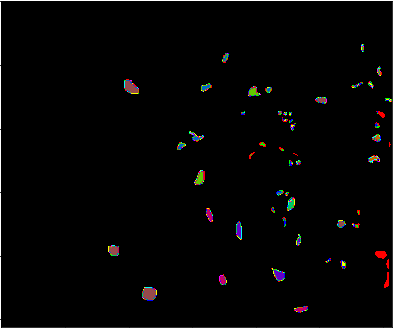}}
\hspace{0.2mm}
\subfloat[]
{\includegraphics[width=0.9in,height=0.9in]{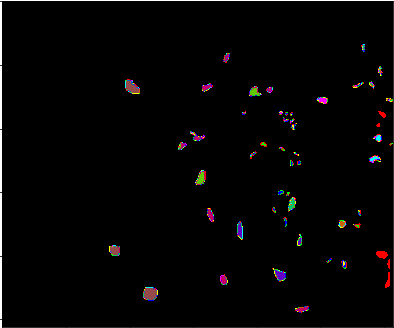}} 
\hspace{0.2mm}
\subfloat[]
{\includegraphics[width=0.9in,height=0.9in]{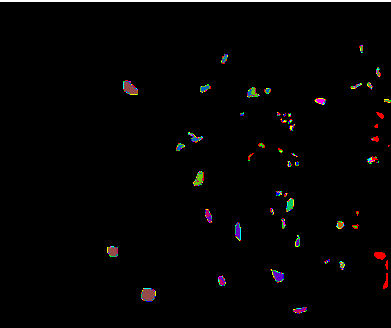}}
\hspace{0.2mm}
\subfloat[]
{\includegraphics[width=0.9in,height=0.9in]{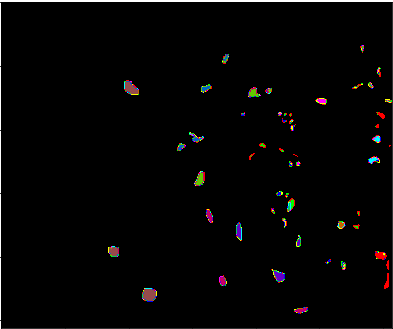}}
\hspace{0.2mm}
\subfloat[]
{\includegraphics[width=0.9in,height=0.9in]{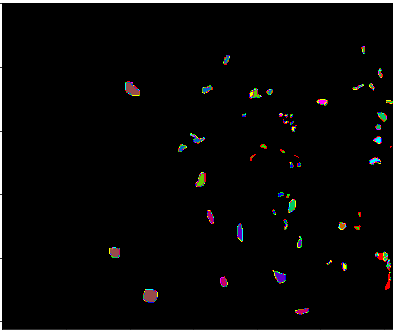}}
\hspace{0.2mm}
\subfloat[]
{\includegraphics[width=0.9in,height=0.9in]{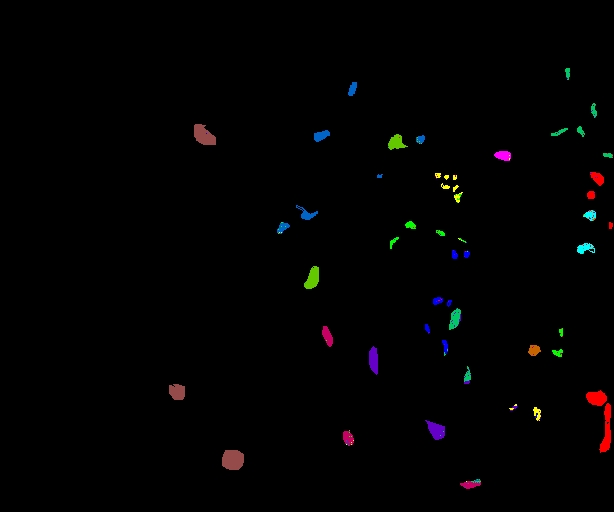}}
\hspace{0.2mm}
\subfloat[]
{\includegraphics[width=0.9in,height=0.9in]{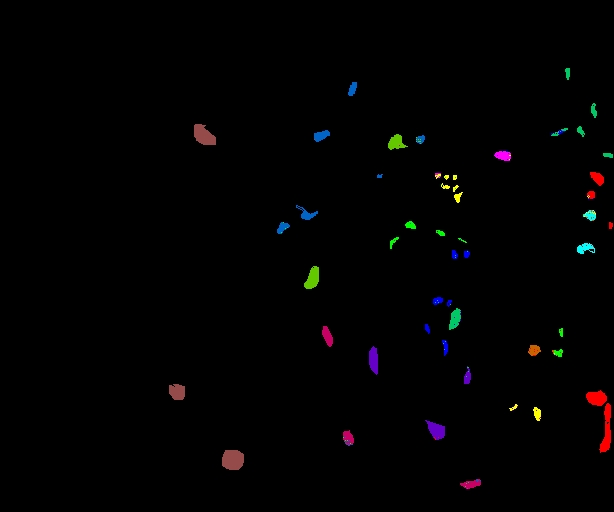}}
\hspace{0.2mm}
\subfloat[]
{\includegraphics[width=0.9in,height=0.9in]{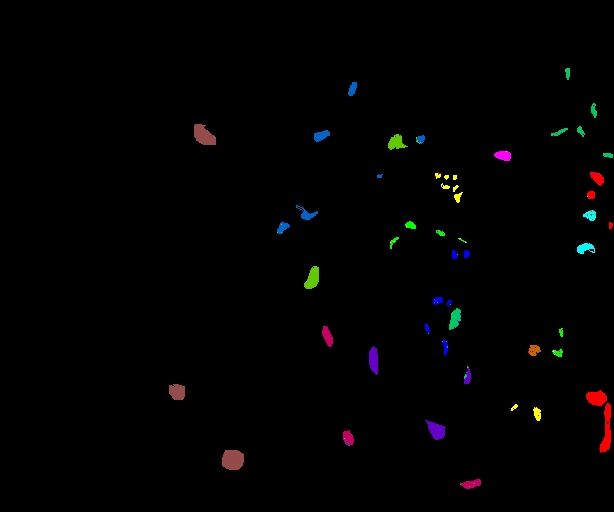}}
\hspace{0.2mm}
\subfloat[]
{\includegraphics[width=0.9in,height=0.9in]{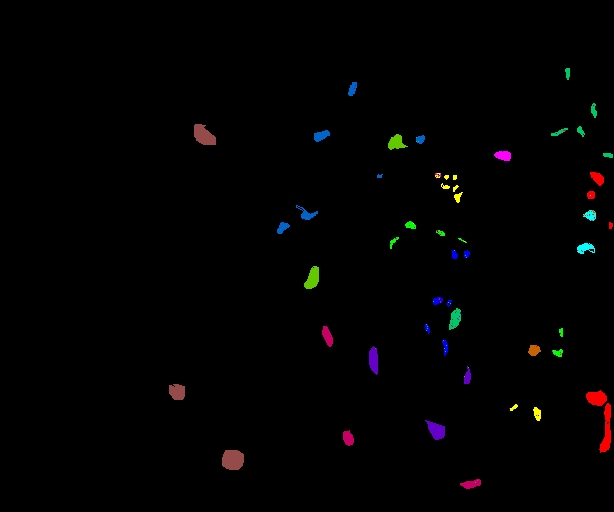}}
\hspace{0.2mm}
\subfloat[]
{\includegraphics[width=0.9in,height=0.9in]{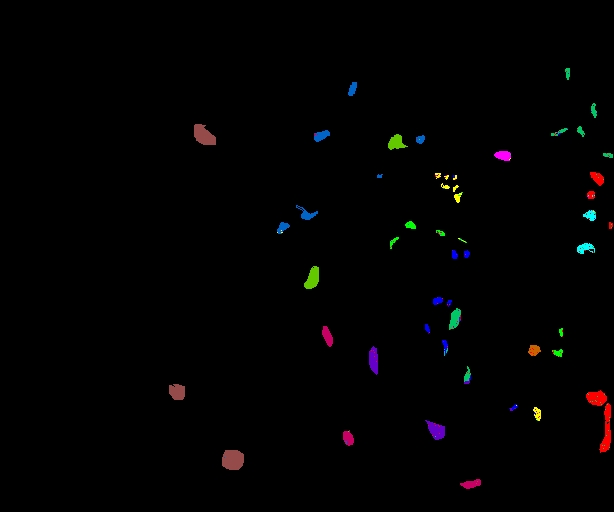}}
\hspace{0.2mm}
\subfloat[]
{\includegraphics[width=0.9in,height=0.9in]{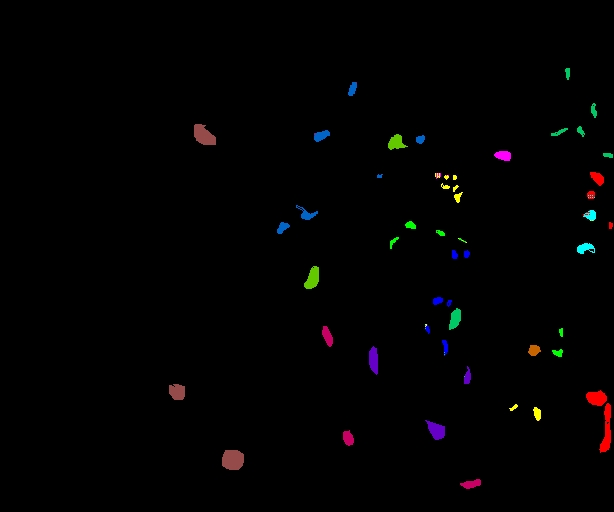}}
\hspace{0.2mm}
\caption{Shows the classification maps of various methods for the KSC data set. (From Left to Right) (a) ground truth, (b) SVM-RBF, (c) CCF-200, (d) 2-D CNN\cite{37ge2020hyperspectral}, (e) GCNN\cite{38hong2020graph}, (f) FADCNN\cite{39yu2021feedback}, (g) NL-GCNN\cite{40mou2020nonlocal}, (h) ViT\cite{46ayas2022spectralswin}, (i) WaveFormer\cite{50waveformer}, (j) PMCN\cite{51PMCN}, (k) PyFormer\cite{48pyramid}, (l) SwinT\cite{47liu2023spectral}, and (m) (Our) DSXFormer.}
\label{fig:8}
\end{figure*}

\subsection{Experimental Setup}
To rigorously evaluate the effectiveness of the proposed DSXFormer framework, extensive comparative experiments were conducted against a diverse set of state-of-the-art (SOTA) HSIC methods, including SwinT \cite{47liu2023spectral}, PyFormer \cite{48pyramid}, PMCN \cite{51PMCN}, WaveFormer \cite{50waveformer}, and ViT \cite{46ayas2022spectralswin}. In addition, several representative traditional and convolution-based approaches were considered for comparison, including SVM-RBF\footnote{https://www.csie.ntu.edu.tw/cjlin/libsvm/}, CCF-200\footnote{https://github.com/twgr/ccfs}, 2D CNN \cite{37ge2020hyperspectral}, GCNN \cite{38hong2020graph}, FADCNN \cite{39yu2021feedback}, and NL-GCNN \cite{40mou2020nonlocal}. For the purpose of reproducibility, the implementation code will be publicly released upon acceptance of this paper\footnote{https://github.com/farhanmarwat}. All experiments were conducted on a workstation equipped with an NVIDIA RTX 4070 GPU and 64 GB of system memory to meet the computational demands of the evaluated models.

The proposed model was implemented using the TensorFlow framework, with an input tensor of size \(25 \times 25 \times 30\), corresponding to the spatial patch dimensions and the number of principal component analysis (PCA) bands retained from the hyperspectral data. The network is designed for multiclass classification and produces class probability outputs through task-specific output layers containing 16, 13, and 9 units, corresponding to the number of categories in the evaluated datasets. A patch size of \(2 \times 2\) is adopted, and a dropout rate of 0.03 is applied to alleviate overfitting. The principal training hyperparameters include a learning rate of \(1 \times 10^{-3}\), a batch size of 128, and a validation split of 10\%. Furthermore, the model employs 8 attention heads, an embedding dimension of 64, and an MLP hidden dimension of 256. Optimization is performed using the AdamW optimizer with a weight decay coefficient of \(1 \times 10^{-4}\), while the loss function is defined as categorical cross-entropy with a label smoothing factor of 0.1. The training procedure is carried out for 100 epochs to ensure stable convergence.

The evaluation metrics adopted in this paper are selected to provide a comprehensive and quantitative assessment of classification performance across different HSIC methods. Detailed definitions and formulations of these metrics are as follow:

\begin{enumerate}
\item \textbf{Overall Accuracy (OA):}  
Overall Accuracy is a primary evaluation metric for assessing the classification performance of HSIC methods. It is defined as the proportion of correctly classified test samples relative to the total number of test samples, thereby providing a direct measure of the global classification effectiveness of a given approach.

\item \textbf{Average Accuracy (AA):}  
Average Accuracy evaluates classification performance at the class level. It is computed by first calculating the individual classification accuracy for each class and subsequently averaging these values across all classes. As such, AA reflects the robustness and balance of the model by highlighting its ability to consistently discriminate among different land-cover categories.

\item \textbf{Kappa Coefficient (K):}  
The Kappa coefficient is a widely adopted statistical metric in HSIC for measuring the agreement between predicted and ground-truth class labels while accounting for chance-level agreement. By quantifying the degree of concordance beyond random classification, the Kappa coefficient provides a reliable and comprehensive assessment of the classification reliability and overall prediction consistency.
\end{enumerate}

\subsection{EXPERIMENTAL RESULTS AND DISCUSSION}

We evaluate the performance of our proposed DSXFormer model on four benchmark HSI datasets: SA, IP, PU, and KSC. These datasets are widely adopted in the HSI classification literature due to their varying spatial resolutions, spectral characteristics, and class imbalances. The SA dataset (512×217 pixels, 204 bands, 16 classes) captures agricultural scenes with large homogeneous regions. The IP dataset (145×145 pixels, 200 bands, 16 classes) is challenging due to severe class imbalance and mixed pixels. The PU dataset (610×340 pixels, 103 bands, 9 classes) represents an urban environment with complex man-made structures. The KSC dataset (512×614 pixels, 176 bands, 13 classes) features wetland vegetation with subtle spectral differences.

DSXFormer is compared against a range of SOTA methods, including classical approaches (SVM-RBF, CCF-200), CNN-based models (2D CNN~\cite{37ge2020hyperspectral}, FADCNN~\cite{39yu2021feedback}), graph convolutional networks (GCNN~\cite{38hong2020graph}, NL-GCNN~\cite{40mou2020nonlocal}), and recent transformer-based architectures (ViT~\cite{46ayas2022spectralswin}, WaveFormer~\cite{50waveformer}, PMCN~\cite{51PMCN}, PyFormer~\cite{48pyramid}, SwinT~\cite{47liu2023spectral}). Performance is reported using Overall Accuracy (OA), Average Accuracy (AA), and Kappa coefficient (K). All experiments follow standard protocols with limited labeled samples.

As shown in Table~\ref{tab:table 6-sa}, on the SA dataset, DSXFormer achieves the highest OA (99.95\%), AA (99.93\%), and K (99.95\%), outperforming the strongest competitor, SwinT (OA: 99.88\%, AA: 99.83\%, K: 99.87\%) by 0.07\% in OA and 0.10\% in AA. This near-perfect performance demonstrates DSXFormer's exceptional ability to exploit both spectral and spatial information in agricultural scenes. Class-wise, it obtains perfect 100\% accuracy in 10 classes, including difficult ones such as Grapes\_untrained (100\% vs. 99.95\% for SwinT) and Vinyard\_untrained (99.96\% vs. 99.88\% for SwinT), which suffer from high intra-class variability. Substantial gains over traditional and CNN-based methods further highlight the effectiveness of its dynamic cross-attention mechanism. 

On the IP dataset, as illustrated in Table~\ref{tab:table7-IP}), which is particularly difficult due to limited samples and spectral mixing, DSXFormer attains the best OA (98.91\%), AA (96.30\%), and K (98.76\%), improving upon PyFormer (OA: 96.67\%) by 2.24\% and PMCN (OA: 96.45\%) by 2.46\%. The high AA reflects robustness to class imbalance. It achieves 100\% accuracy in six classes and shows notable improvements in large soybean classes (Soybean-mintill: 99.42\%; Soybean-clean: 99.04\%) as well as competitive performance on the extremely small Oats class (64.29\%). These results underscore DSXFormer's ability to extract discriminative features even with scarce labeled data.

For the PU dataset, as show in the Table~\ref{tab:table8-PU}, DSXFormer delivers SOTA OA (99.85\%), AA (99.70\%), and K (99.80\%), surpassing SwinT (OA: 99.41\%) by 0.44\%. It achieves perfect or near-perfect accuracy in eight of nine classes, with significant gains in Gravel (99.73\%) and Bitumen (99.89\%), demonstrating superior handling of urban material confusion and sharp boundaries. On the KSC dataset (Table~\ref{tab:table 9-ksc}, DSXFormer records the top OA (98.52\%), AA (97.51\%), and K (98.35\%), outperforming PyFormer (OA: 97.34\%) by 1.18\%. It reaches 100\% in six classes, with strong gains in marsh-related classes exhibiting subtle spectral differences influenced by water content, confirming the effectiveness of its adaptive spectral processing in ecologically diverse scenes.

Across all four datasets, DSXFormer consistently achieves the highest OA, AA, and K, with average OA gains of approximately 1.0–2.5\% over the second-best transformer-based method. It excels particularly on challenging classes with high intra-class variance or limited samples, attributable to its dynamic spectral cross-attention and dual-branch architecture, which adaptively model band interactions and spatial contexts more effectively than fixed-window transformers or pyramidal designs. Traditional and CNN-based methods lag due to inadequate global modeling, while graph networks struggle with long-range dependencies. The consistently high AA indicates balanced performance, critical for real-world applications with imbalanced land cover. Minor limitations in extremely small classes suggest potential for further few-shot enhancements. Overall, the results firmly establish DSXFormer as a new SOTA for HSIC.

\section{Conclusion and Future Work}
In this paper, we propose DSXFormer, a novel DSXFormer for HSIC. The proposed framework integrates a DSX mechanism with a transformer architecture to jointly model fine-grained spectral dependencies and long-range spatial–spectral contextual information. By exploiting complementary global and local spectral statistics, the DSX block adaptively highlights informative spectral bands while suppressing redundant responses, resulting in more discriminative and compact feature representations. In addition, the hierarchical transformer structure enables efficient multi-scale contextual aggregation with limited computational overhead. Extensive experiments on four benchmark datasets, such as SA, IP, PU, and KSC, demonstrate the consistent superiority of DSXFormer over classical machine learning methods, CNN-based models, and SOTA transformer-based approaches. DSXFormer achieves SOTA OA of 99.95\% (SA), 98.91\% (IP), 99.85\% (PU), and 98.52\% (KSC), along with high average accuracy and Kappa coefficients, validating its robustness and generalization capability across diverse hyperspectral scenes with spectral redundancy and limited labeled samples.

Future work will focus on further enhancing the flexibility and scalability of the proposed DSXFormer framework. Although DSXFormer effectively balances spectral discrimination and contextual modeling, its reliance on fixed patch-based tokenization may limit adaptability in highly heterogeneous scenes. To address this, future work will investigate adaptive tokenization schemes, dynamic multi-scale attention mechanisms, and enhanced spectral–spatial interaction modules. In addition, extending DSXFormer to semi-supervised, few-shot, and self-supervised learning paradigms will be explored to reduce the dependence on large amounts of labeled data. Another promising direction involves incorporating temporal–spectral modeling to enable effective analysis of time-varying hyperspectral data in dynamic environments.


\bibliographystyle{IEEEtran}

\end{document}